%% file: main.tex
\documentclass[10pt,twocolumn,letterpaper]{article}

\usepackage{soul}

\usepackage{comment}
\usepackage[pagenumbers]{cvpr} %

\input{preamble}

\def\paperID{7213} %
\def\confName{CVPR}
\def\confYear{2024}

\newcommand{\method}{ABNN}
\newcommand{\methods}{ABNNs}

\title{Make Me a BNN: A Simple Strategy for Estimating\\
Bayesian Uncertainty from Pre-trained Models}

\author{\textbf{Gianni Franchi},\textsuperscript{\rm 1, *, $\dagger$}  \textbf{Olivier Laurent},\textsuperscript{\rm 1, 2, *}  \textbf{Maxence Leguéry},\textsuperscript{\rm 1} 
\textbf{Andrei Bursuc},\textsuperscript{\rm 3} \\
\textbf{Andrea Pilzer}\textsuperscript{\rm 4} \& \textbf{Angela Yao}\textsuperscript{\rm 5} \\
 U2IS, ENSTA Paris, Institut Polytechnique de Paris,\textsuperscript{\rm 1} Université Paris-Saclay,\textsuperscript{\rm 2}  \\
NVIDIA,\textsuperscript{\rm 3} valeo.ai,\textsuperscript{\rm 4} National University of Singapore\textsuperscript{\rm 5}
}

\input{abbrev.tex}

\begin{document}
\twocolumn[{
\renewcommand\twocolumn[1][]{#1}%
\maketitle
\begin{center}
    \centering
    \captionsetup{type=figure}
    \captionsetup[subfigure]{labelformat=empty}
     \addtocounter{figure}{-1}
     \begin{subfigure}[b]{0.49\textwidth}
         \centering
         \includegraphics[width=0.86\textwidth]{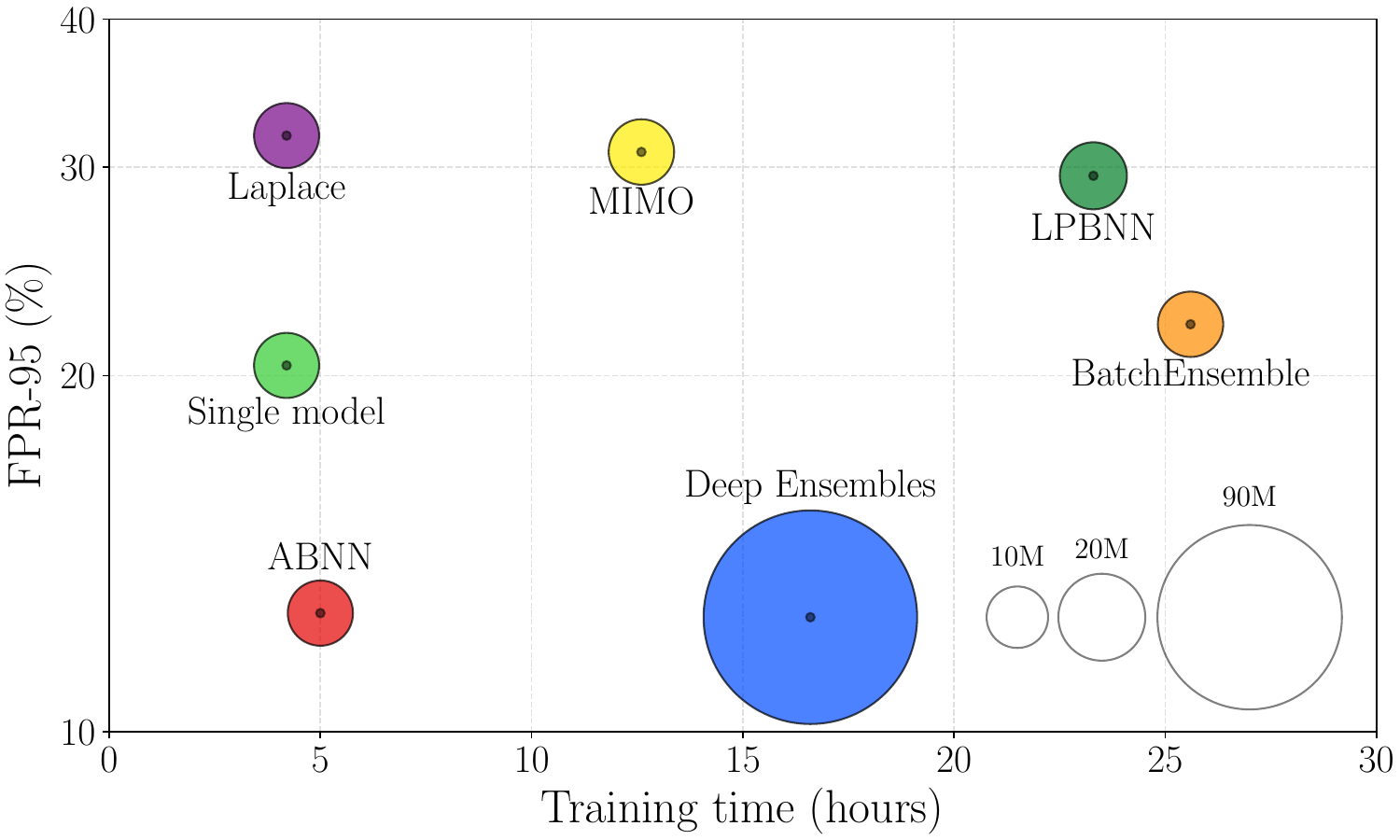}
         \label{figteaser:Label-anom}
     \end{subfigure}
          \hfill
     \begin{subfigure}[b]{0.49\textwidth}
         \centering
         \includegraphics[width=0.86\textwidth]{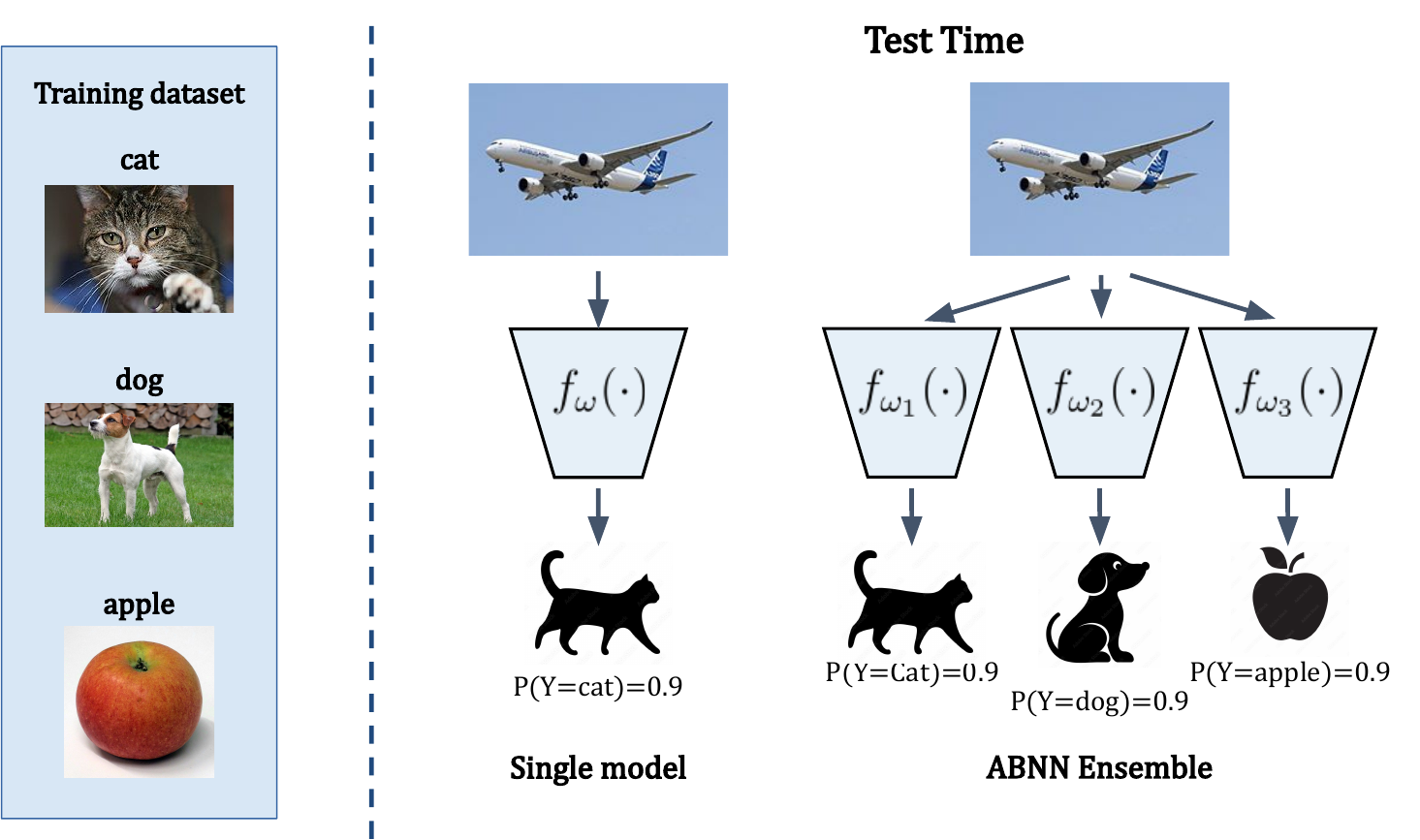}
         \label{figteaser:RobustaGeneration}
         \vspace{5pt}
     \end{subfigure}
    \captionof{figure}{\textbf{Benefits of the ABNN approach.} \textbf{(left)} Evaluation of the trade-off between computational cost (training time and model size) and performance (in terms of FPR95 score, lower the better) for various uncertainty quantification techniques on CIFAR-10~\cite{krizhevsky2009learning} with WideResNet~\cite{zagoruyko2016wide} and ensembles of size 4. \textbf{(right)} Using ABNN ensembling at test time: given an out-of-distribution input, a simple DNN may make a high-confidence incorrect predictions, whereas ABNN produces more uncertain decisions through its diverse predictions.}
    \label{fig:qualitative_robustness}
\end{center}}]

\def\thefootnote{*}
\begin{NoHyper}
\footnotetext{equal contribution, \quad \textsuperscript{$\dagger$} {\tt gianni.franchi@ensta-paris.fr}}
\end{NoHyper}
\def\thefootnote{\arabic{footnote}}
\input{sec/0_abstract}    
\input{sec/1_intro_related}

\input{sec/2_contribution}

\input{sec/3_Experiments}

\clearpage
{
    \small
    \bibliographystyle{ieeenat_fullname}
    \bibliography{main}
}

\appendix
\input{sec/X_suppl}

\end{document}

%% file: preamble.tex
\usepackage[dvipsnames]{xcolor}
\usepackage{booktabs}
\usepackage{colortbl}
\usepackage{multirow}

\newcommand{\second}{\cellcolor{blue!10}}
\newcommand{\first}{\cellcolor{blue!30}}

\usepackage{dsfont}
\usepackage{algorithm}
\usepackage{algpseudocode}
\usepackage{algpascal}
\usepackage{graphbox}
\usepackage{color}
\usepackage[dvipsnames]{xcolor}
\usepackage{soul}
\usepackage{titletoc}

\usepackage{amsthm}

\definecolor{cvprblue}{rgb}{0.21,0.49,0.74}
\usepackage[pagebackref,breaklinks,colorlinks,citecolor=cvprblue]{hyperref}

%% file: abbrev.tex
\makeatletter
\DeclareRobustCommand\onedot{\futurelet\@let@token\@onedot}
\def\@onedot{\ifx\@let@token.\else.\null\fi\xspace}

\makeatother

\definecolor{mygray}{gray}{0.5}
\newcommand{\gray}[1]{\textcolor{mygray}{#1}}
\newcommand{\purple}[1]{\textcolor{DarkOrchid}{#1}}

\newcommand{\vomega}{\boldsymbol{\mathbf{\omega}}}
\newcommand{\vepsilon}{\boldsymbol{\mathbf{\epsilon}}}

\newcommand{\vy}{\mathbf{y}}
\newcommand{\vh}{\mathbf{h}}
\newcommand{\va}{\mathbf{a}}
\newcommand{\vx}{\mathbf{x}}
\newcommand{\vu}{\mathbf{u}}

\DeclareMathOperator{\soft}{soft}
\DeclareMathOperator{\normop}{norm}

\DeclareMathOperator{\BNL}{\textbf{BNL}}

\usepackage{pifont}
\definecolor{dgreen}{rgb}{0.0, 0.5, 0.0}
\definecolor{better}{rgb}{0.19, 0.55, 0.91}
\definecolor{worse}{rgb}{0.82, 0.1, 0.26}
\newcommand{\cmark}{\textcolor{dgreen}{\ding{51}}}
\newcommand{\xmark}{\textcolor{worse}{\ding{55}}}

%% file: sec/0_abstract.tex
\begin{abstract}
\vspace{-5pt}
Deep Neural Networks (DNNs) are powerful tools for various computer vision tasks, yet they often struggle with reliable uncertainty quantification — a critical requirement for real-world applications. Bayesian Neural Networks (BNN) are equipped for uncertainty estimation but cannot scale to large DNNs that are highly unstable to train.
To address this challenge, we introduce the Adaptable Bayesian Neural Network (\method{}), a simple and scalable strategy to seamlessly transform DNNs into BNNs in a post-hoc manner with minimal computational and training overheads.
\method{} preserves the main predictive properties of DNNs while enhancing their uncertainty quantification abilities through simple BNN adaptation layers (attached to normalization layers) and a few fine-tuning steps on pre-trained models. We conduct extensive experiments across multiple datasets for image classification and semantic segmentation tasks, and our results demonstrate that \method{} achieves state-of-the-art performance without the computational budget typically associated with ensemble methods.

\end{abstract}

%% file: sec/1_intro_related.tex
\vspace{-5pt}
\section{Introduction}
\label{sec:intro}

Deep Neural Networks (DNNs) have emerged as powerful tools with a profound impact on various perception tasks, such as image classification~\cite{alexnet2012neurips, dosovitskiy2020image}, object detection~\cite{he2017mask, redmon2016you}, natural language processing~\cite{severyn2015,devlin2018bert,radford2018improving}, etc. 
With this progress, there is growing excitement and expectation about the potential applications of  DNNs across industries.
To meet this end, there is a critical need to address a fundamental challenge: improving DNN reliability by quantifying the inherent uncertainty in their predictions~\cite{tran2022plex,hendrycks2021unsolved,hendrycks2021jacob}. Deploying DNNs in real-world applications, particularly in safety-critical domains such as autonomous driving, medical diagnoses, industrial visual inspection, etc., requires a comprehensive understanding of their limitations and vulnerabilities beyond their raw predictive accuracy, often considered a primary performance metric. 
By quantifying the uncertainty within these models with millions of parameters and non-trivial inner-working~\cite{arrieta2020explainable, zablocki2022explainability} and failure modes~\cite{hein2019relu, neuhaus2023spurious} in front of the many different long-tail scenarios~\cite{li2022coda, chan2021segmentmeifyoucan}, we can make informed decisions about when and how to rely on their predictions.

In deep learning, uncertainty estimation has been traditionally addressed under Bayesian approaches drawing inspiration from findings in Bayesian Neural Networks (BNNs)~\cite{mackay1992practical, neal2012bayesian, blundell2015weight, welling2011bayesian} that stand on solid theoretical grounds and properties~\cite{nalisnick2018priors, wilson2020bayesian, izmailov2021bayesian}. BNNs estimate the posterior distribution of the model parameters given the training dataset. Ensembles can be sampled from this distribution at runtime, and their predictions can be averaged for reliable decisions. BNNs promise improved predictions and reliable uncertainty estimates with intuitive decomposition of the uncertainty sources~\cite{depeweg2018decomposition, hullermeier2021aleatoric}. However, although they are easy to formulate, BNNs are notoriously difficult to train over large DNNs~\cite{ovadia2019can,dusenberry2020efficient}, in particular for complex computer vision tasks, 
due to training instability and computational inefficiency as they are typically trained through variational inference~\cite{jordan1999introduction}. This limitation of BNNs has inspired two major lines of research toward scalable uncertainty estimation: ensembles and last-layer uncertainty approaches.

Deep Ensembles~\cite{lakshminarayanan2017simple} emerge as a simple and highly effective alternative to BNNs for uncertainty estimation on large DNNs. Deep Ensembles are trivial to train by essentially instantiating the same training procedure over different weight initializations of the same network and have been shown to preserve many of the properties of BNNs in terms of predictions diversity~\cite{fort2019deep, wilson2020bayesian}. This is a beneficial property for out-of-distribution (OOD) generalization~\cite{nayman2022diverse}. However, their high computational cost (during both training and inference) makes them inapplicable to many practical applications with computational constraints. In the last few years, multiple computationally efficient alternatives have emerged aiming to reduce training cost during training and inference~\cite{maddox2019simple, franchi2020tradi, franchi2020encoding, wen2019batchensemble, havasi2021training, gal2016dropout, daxberger2021bayesian}. However, these methods propose specific network architectures and non-trivial trade-offs in computational cost, accuracy, and predictive uncertainty quality.

Last-layer uncertainty approaches aim for BNNs with fewer stochastic weights to produce ensembles or uncertainty estimates~\cite{malinin2018predictive, laplace2021, brosse2020last,kristiadi2020being,corbiere2021beyond}. These methods leverage popular DNN architectures~\cite{he2016deep} to which they attach a stochastic layer and train all parameters for a complete training cycle. While training stability is improved compared to standard BNNs, joint optimization of deterministic and stochastic parameters requires careful tuning. Daxberger et al.~\cite{laplace2021} propose training the last layer separately in a \emph{post-hoc} manner effectively leveraging Laplace approximation for optimization\cite{tierney1986accurate,mackay1992practical,ritter2018scalable}. Decoupling the optimization of the encoder from the uncertainty layer enables the use of the typical training recipes for the encoder or simply leveraging off-the-shelf pre-trained networks. The limitation of last-layer methods is related to the access to only high-level features for producing uncertainty estimates. In contrast, signals of distribution shift or small anomaly patterns (e.g., in semantic segmentation) can be found primarily on low-level features earlier in the network. Indeed, strong uncertainty estimation methods leverage information from multiple layers of the networks~\cite{lee2018simple, franchi2020encoding, daxberger2021bayesian}.

In this work, we aim for scalable and effective uncertainty estimation without sophisticated optimization schemes and potential training instability and without compromising predictive performance.
We propose a post-hoc strategy that starts from a pre-trained DNN and transforms it into a BNN with a simple plug-in module attached to the normalization layers and only a few epochs of fine-tuning. We show that this strategy, dubbed Adaptable-BNN (\method{}), can estimate the posterior distribution around the local minimum of the pre-trained model in a resource-efficient manner while still achieving competitive uncertainty estimates with diversity. Furthermore, ABNN allows for sequential training of multiple BNNs starting from the same checkpoint, thus modeling various modes within the true posterior distribution.

Our contributions are: \textbf{(1)} We propose ABNN, a simple strategy to transform a pre-trained DNN into a BNN with uncertainty estimation capabilities. ABNN is computationally efficient and compatible with multiple DNN architectures (ConvNets: ResNet-50, WideResnet28-10; ViTs), provided they are equipped with normalization layers. \textbf{(2)} We observe that the variance of the gradient for ABNN's parameters is lower compared to that of a classic BNN, resulting in a more stable backpropagation.
\textbf{(3)} Extensive experiments validate that ABNN, although simple and computationally frugal, achieves competitive performance in terms of accuracy and uncertainty estimation over multiple datasets and tasks: image classification (CIFAR-10, CIFAR-100~\cite{krizhevsky2009learning}, ImageNet~\cite{deng2009imagenet}) and semantic segmentation (StreetHazards, BDD-Anomaly~\cite{hendrycks2019anomalyseg}, MUAD~\cite{franchi2022muad}) in both in- and out-of-distribution settings.

\section{Related work}

\noindent\textbf{Epistemic uncertainty and Bayesian posterior.}
Tackling epistemic uncertainty estimation~\citep{hora1996aleatory} - the uncertainty on the model itself - is essential to improve the reliability of DNNs~\cite{hullermeier2021aleatoric}. However, obtaining satisfying approximations of this uncertainty remains a challenge as it requires a scalable estimation of the extremely high dimensional distribution of the weights, the \emph{posterior}. Our work presents \method, a significantly more scalable method compared to the BNNs~\citep{goan2020bayesian} that predominantly shape the landscape of epistemic uncertainty estimations~\citep{gawlikowski2023survey}.

\noindent\textbf{Bayesian Neural Networks and Ensembles.}
BNNs~\cite{tishby1989consistent} formulate probabilistic predictions by both introducing explicit and controllable prior knowledge on the network weights~\cite{neal2012bayesian, kendall2017uncertainties} and estimating the posterior. While formulating mathematically, the posterior distribution is possible~\cite{wilson2020bayesian}, its computation for modern models is intractable. This need for scalability leads to approximation techniques that include variational inference~\cite{blundell2015weight, blei2017variational} BNNs, which fit simpler distributions to the posterior (diagonal Gaussian for the former). Many other approximation methods have been proposed, like the efficient probabilistic backpropagation~\cite{hernandez2015probabilistic}, Monte-Carlo Dropout~\cite{gal2016dropout} that model the posterior as a mixture of Diracs and Laplace methods that estimate the posterior thanks to the local curvature of the loss~\cite{tierney1986accurate,mackay1992practical,ritter2018scalable, subhankar2022uncertainty}. However, \emph{deep ensembles}~\cite{hansen1990neural, lakshminarayanan2017simple}, the most successful solution is simpler. It consists of simply averaging the predictions of several independently trained models. This method is powerful as it improves the reliability and the quality of the predictions, albeit costly in training and inference. Many approximate methods stepped into the breach and proposed to reduce the number of parameters, the training time, or the number of forward passes~\cite{wen2019batchensemble, franchi2020encoding, franchi2020tradi, havasi2021training, laurent2023packed, xia2023window}. The proposed method is in this line of scalability and computational efficiency.

\noindent\textbf{Post-hoc uncertainty quantification.}
In today's context of computer vision, with ever-increasing datasets~\cite{schuhmann2022laion, kirillov2023segment}, model sizes~\cite{dehghani2023scaling}, and inference constraints, post-hoc methods could be a solution to benefit from both the power of foundation models~\cite{ilharco2021openclip, rombach2022high,zhai2023sigmoid} and uncertainty quantification. The Laplace method reigns among these post-hoc uncertainty quantification methods, especially with its last-layer approximations~\cite{ritter2018scalable} that make them even more scalable. However, even with the coarse approximation of the posterior achieved by Kronecker-factored Laplace~\cite{subhankar2022uncertainty}, it is not always very efficient with modern datasets. We propose a straightforward and effective method that works with pre-trained models.

\label{sec:Related}

%% file: sec/2_contribution.tex
\section{Background}\label{section:background}

We start by introducing our formalism and a brief overview of the Bayesian posterior and BNNs for uncertainty quantification.

\subsection{Preliminaries}

\noindent \textbf{Notations.}
Let us denote $\mathcal{D}=\{(\vx_i, y_i)\}_{i=1}^{N}$ the training set containing $N$ samples and labels drawn from a joint distribution $P_{(X, Y)}$. The input $\vx_i\in\mathbb{R}^d$ is processed by a neural network $f_{\vomega}$, of parameters $\vomega$, that outputs classification predictions $\hat{y}_i= f_{\vomega}(\vx_i)\in\mathbb{R}$.

\noindent\textbf{From MLE to MAP.} 
In our context, $P(Y\!=\!y_i \mid X\!=\!\vx_i, \vomega)$ is a categorical distribution over the classes within the range of $Y$. %
We omit the random variable notation in the following for clarity.
The log-likelihood of this distribution typically corresponds to the cross-entropy loss, which practitioners often minimize with stochastic gradient descent to obtain a maximum likelihood estimate (MLE):
$\mathcal{L}_{\scriptscriptstyle \text{MLE}}(\vomega) = - \sum_{(\vx_i,y_i) \in \mathcal{D}} \log P(y_i\mid \vx_i,\vomega)$.

Going further, the Bayesian framework allows us to incorporate prior knowledge regarding $\vomega$ denoted as the distribution %
$P(\vomega)$ that complements the likelihood and leads to the research of the maximum a posteriori (MAP),
via the minimization of the following loss function:
\begin{align}\label{eq:loss_MAP}
\mathcal{L}_{\scriptscriptstyle \textrm{MAP}}(\vomega)
  = - \sum_{(\vx_i,y_i) \in \mathcal{D}} \log P(y_i \mid \vx_i,\vomega) - \log P(\vomega).
\end{align}
The normal prior is the standard choice for $P(\vomega)$, leading to the omnipresent L2 weight regularization.

\subsection{Bayesian Posterior and BNNs}
Typically, DNNs retain a single set of weights $\vomega_\textrm{MAP}$ at the end of the training to use at inference. As such, we de facto consider this model as an oracle. In contrast, BNNs attempt to model the posterior distribution $P(\vomega \mid \mathcal{D})$ to take all possible models into account. The prediction $y$ for a new sample $\vx$ is computed as the expected outcome from an infinite ensemble, including all possible weights sampled from the posterior distribution:
\begin{equation}\label{eq:marginalization}
P(y \mid \vx,\mathcal{D})  = \int\limits_{\vomega\in\Omega} P(y \mid \vx, \vomega) P(\vomega \mid \mathcal{D}) d\vomega.
\end{equation}
However, in practice, this Bayes ensemble approach is intractable since the integral Eq.~\eqref{eq:marginalization} is computed over the entire parameter space $\Omega$. 
Practitioners~\cite{lakshminarayanan2017simple} approximate this integral by averaging predictions derived from a finite set $\{\vomega_1, \ldots \vomega_M\}$ of $M$ weight configurations sampled from the posterior distribution: 
\begin{eqnarray}\label{eq:approximate_likelihood}
P(y \mid \vx, \mathcal{D}) \approx \frac{1}{M} \sum_{m=1}^M  P( y \mid \vx, \vomega_m).
\end{eqnarray}

Let us start with a simple Multi-Layer Perceptron (MLP) with two hidden layers without loss of generality. For a given input data point $\vx$, the prediction of the DNN is defined as, with $\vh_1$ and $\vh_2$  the preactivation maps, $a(\cdot)$ the activation function:
\begin{align}\label{eq:DNN1}
\begin{split}
&\vh_1 = W^{(1)}\vx \\
&\vu_1 = \normop(\vh_1,\beta_1,\gamma_1)= \frac{\vh_1 - \hat{\mu}_{1}}{\hat{\sigma}_{1}}\times\gamma_1 +\beta_1 \\
&\va_1 = a(\vu_1) \\
&\vu_2 = \normop\left(W^{(2)}\va_1,\beta_2,\gamma_2\right) \mbox{, and } \va_2 = a(\vu_2) \\
&\vh_3 = W^{(3)}\va_2 \mbox{, and } P(y \mid \vx,  \vomega) = \soft(\vh_3)
\end{split}
\end{align}
In Eq. \eqref{eq:DNN1}, $\soft(\cdot)$ is the softmax, $\va_1$ and $\va_2$ are the hidden activations, and $\{W^{(j)}\}_{j\in\{0, 1, 2\}}$ correspond to the weights of the linear layers. The operator $\normop(\cdot, \beta_j, \gamma_j)$, of trainable parameters $\beta_j$ and $\gamma_j$, can refer to any batch, layer, or instance normalization (BN, LN, IN). Finally,  $\normop$ comes with its empirical mean $\hat{\mu}_{\vu_j}$ and variance $\hat{\sigma}_{\vu_j}$. We omit the small value often added for computational stability.

In the current form, we can leverage this architecture to learn different tasks. However, modeling the uncertainty of the predictions beyond the use of softmax scores as confidence proxies is non-trivial. 
BNNs~\cite{blundell2015weight} are one solution to improve uncertainty estimation. Generally, they hypothesize the independence of the layers and sample from the resulting posterior estimate. For the $j$-th layer, this yields:
\begin{align}\label{eq:BNN1}
\begin{split}
&\vu_j = \normop(W^{(j)}\vx,\beta_j,\gamma_j) \mbox{, } W^{(j)} \sim P(W^{(j)}|\mathcal{D})\mbox{, and} \\
&\va_j = a(\vu_j). \\
\end{split}
\end{align}
As such, BNNs approximate the marginalization~\eqref{eq:marginalization} of the parameters - an extremely complex task - by generating multiple predictions. Variational inference BNNs~\cite{blundell2015weight}, the most scalable version among these methods, base their estimation on the "reparametrization trick", here at layer $j$:
\begin{align}\label{eq:BNN1-2}
\begin{split}
&\vu_j = \normop\left(\left[ W^{(j)}_{\mu}  + \vepsilon_{j} W^{(j)}_{\sigma} \right]\vh_{j-1},\beta_j,\gamma_j\right) \mbox{ and}\\
&\va_j = a(\vu_j),
\end{split}
\end{align}
where the matrices $W^{(j)}_{\mu}$ and $W^{(j)}_{\sigma}$ denote the mean and standard deviation of the posterior distribution of layer $j$, and $\vepsilon_{j} \sim \mathcal{N}(\mathbf{0}, \mathds{1})$ is a zero-mean unit-diagonal Gaussian vector or matrix. %
This method enables learning an estimate of a diagonal posterior distribution at the cost of tripling the number of parameters compared to a standard network.

\section{\method{}}
\label{sec:formatting}
\begin{figure*}[t]
    \centering{\includegraphics[width=0.92\linewidth]{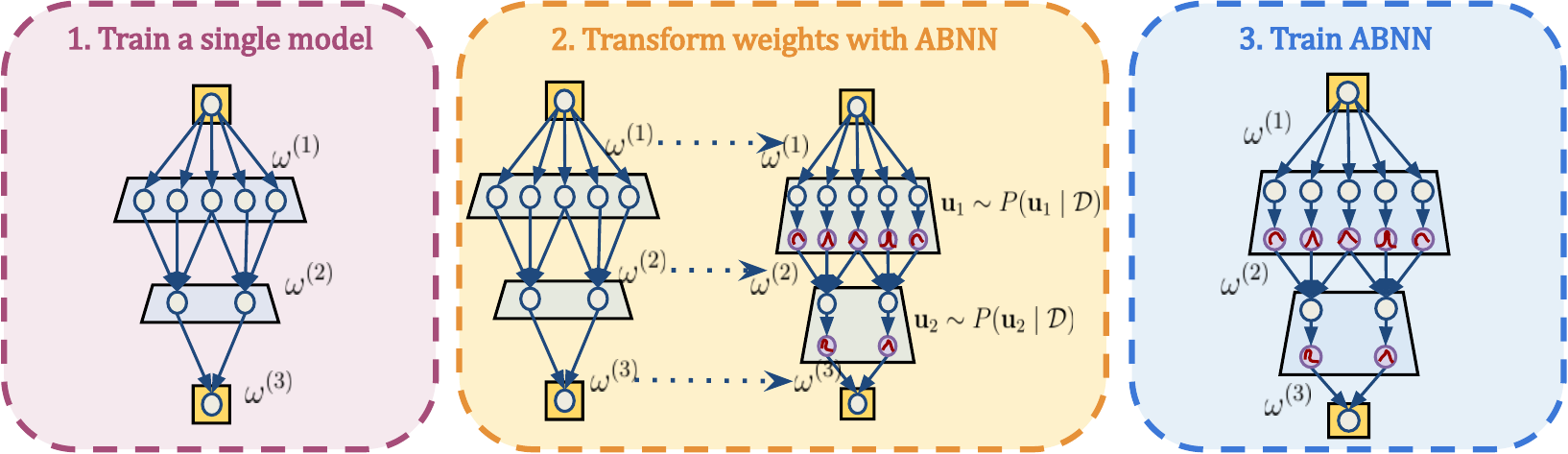}}
    \caption{\textbf{Illustration of the training process for the \method{}.} The procedure begins with training a single DNN $\vomega_{\textrm{MAP}}$, followed by architectural adjustments to transform it into an \method{}. The final step involves fine-tuning the \method{} model.}
    \label{fig:training_process}
\end{figure*}

\subsection{Converting DNNs into BNNs}

We base our post-hoc Bayesian strategy on 
pre-trained DNNs that incorporate normalization layers such as batch %
~\cite{ioffe2015batch}, layer%
~\cite{ba2016layer}, or instance normalization~\cite{ulyanov2016instance}. This is not a limiting factor as most modern architectures include one type of these layers~\cite{he2016deep, dosovitskiy2020image, liu2022convnet}. Subsequently, we modify these normalization layers by introducing a Gaussian perturbation, incorporating our novel Bayesian Normalization Layer (BNL). This adaptation aims to transform the initially deterministic DNN into a BNN. The introduction of the BNL allows us to efficiently leverage pre-trained models, facilitating the conversion to a BNN with minimal alterations.
We propose replacing the normalization layers with our novel Bayesian normalization layers (BNL) that incorporate Gaussian noise to transform the deterministic DNNs into BNNs easily. BNLs unlock the power of pre-trained models for uncertainty-aware Bayesian networks.
Formally, our BNN is defined as:
\begin{align}\label{eq:BNL1}
\begin{split}
&\vu_j = \BNL\left(W^{(j)}\vh_{j-1}\right),\mbox{ and} \\
&\va_j = a(\vu_j)\mbox{ with} \\
&\BNL(\vh_j)  = \frac{\vh_j - \hat{\mu}_{j}}{\hat{\sigma}_{j} }\times\gamma_j (1+\vepsilon_{j})+\beta_{j}.\\
\end{split}
\end{align}
The empirical mean and variance are still represented by  $\hat{\mu}_{\vu_j}$ and $\hat{\sigma}_{\vu_j}$ and computed through batch, layer or instance normalization. In the equation, $\vepsilon_{j} \sim \mathcal{N}(\mathbf{0}, \mathds{1})$ signifies a sample drawn from a Normal distribution, and $\gamma_{j}$ and $\beta_{j}$ are the two learnable vectors of the $j$-th layer.

The DNN being transformed into a BNN, we exclusively retrain the parameters  $\gamma_j$ and $\beta_j$ for a limited number of epochs using the loss introduced in Section \ref{sec:loss}. To further improve its reliability and generalization properties~\cite{wilson2020bayesian}, we do not train a singular \method{}, but rather multiple copies of \methods{}, as explained in section \ref{sec:loss}, resulting in a finite set ${\vomega_1, \ldots \vomega_M}$ of $M$ weight configurations. We discuss the benefits of this multi-modality in Appendix \ref{sup:Multimodes}.

During inference, for each sample from  ABNN $\vomega_m$, we augment the number of samples by independently sampling multiple $\vepsilon_{j} \sim \mathcal{N}(\mathbf{0}, \mathds{1})$. With $\vepsilon$ the concatenation of all $\vepsilon_{j}$, and $\{ \vepsilon_{l} \}_{l\in[1, L]}$ the set of $\vepsilon$s, each individual \method{} sample is expressed as $P(y\mid x,\omega,\vepsilon)$. During inference, the prediction $y$ for a new sample $\vx$ is computed as the expected outcome from a finite ensemble, encompassing all the weights sampled from the posterior distribution:
\begin{align}\label{eq:marginalization_ABNN}
P(y \mid \vx, \mathcal{D}) \approx \frac{1}{ML} \sum_{l=1}^L \sum_{m=1}^M  P( y \mid \vx, \vomega_m, \vepsilon_{l}).
\end{align}
\subsection{\method{} training loss}\label{sec:loss}
The multimodality~\cite{fort2019deep,izmailov2021bayesian,laurent2023symmetry} of the posterior distribution of DNNs is a challenge to any attempt to perform variational inference with on a mono-modal distribution.
Wilson and Izmailov~\cite{wilson2020bayesian} have proposed to tackle this issue by training multiple BNNs. However, such approaches inherit the instability of classical BNNs and may struggle to capture different modes accurately.
\method{} encounters a similar challenge, requiring safeguards against collapsing into the same local minima during post-training. To mitigate this problem, we introduce a small perturbation to the loss function, preventing collapse and encouraging diversity into the training process. This perturbation involves a modification of the class weights within the cross-entropy loss, now defined as:
\begin{align}\label{eq:loss_perturb}
\mathcal{E}(\vomega) = -\hspace{-0.3cm}\sum_{(\vx_i,y_i) \in \mathcal{D}} \hspace{-0.3cm}\eta_i \log P(y_i \mid \vx_i,\vomega).
\end{align}
In this formula, $\eta_i$ represents the class-dependent random weight we initialize at the beginning of training. Typically, it can be set to zero or one to amplify the effect of certain classes.
In contrast to classical variational BNNs~\cite{blundell2015weight} that optimize the evidence lower-bound loss, \method{} maximizes the MAP. The optimization involves the following loss:
\begin{equation}\label{eq:loss_MAP_multi1}
\mathcal{L}(\vomega)
  =\mathcal{L}_{\scriptscriptstyle \textrm{MAP}}(\vomega) + \mathcal{E}(\vomega).
\end{equation}
We employ this procedure to train various ABNNs, resulting in ABNNs trained with different losses due to the presence of $\mathcal{E}(\vomega)$.

\subsection{\method{} training procedure}  

\method{} is trained through a post-hoc process designed to leverage the strength of Bayesian concepts and improve the uncertainty prediction of DNNs. The training pseudo code for ABNN, detailed in Alg. \ref{algopseudo}, outlines the step-by-step process of transforming a conventional DNN into an ensemble of Bayesian models. 

We start from a pre-trained neural network (Alg. \ref{algopseudo}, line 1) and introduce the Bayesian normalization layers, replacing the old batch, instance or layer normalization of the former DNN (Alg. \ref{algopseudo} lines 2 to 10). This operation transforms the conventional deterministic network into a BNN to help quantify the uncertainty. We initialize the weights of the new layer with the values of the replaced normalizations.

Then, we fine-tune the modified network to capture better the inherent uncertainty (Alg. \ref{algopseudo} lines 12 to 24). The full process is described in Figure~\ref{fig:loss_posterior}, providing a clear overview of the modifications made to enable Bayesian modeling. More precisely, we only fine-tune the normalization weights in Table~\ref{tab:classif}.
To improve the posterior estimation of our ABNN models, we fine-tune multiple instances of the normalization layers (typically 3 to 4).  This ensemble approach provides robustness and contributes to a more reliable estimation of the posterior distribution. Training multiple ABNNs, each starting from the same checkpoint, enhances our ability to capture diverse modes of the true posterior, thereby improving the overall uncertainty quantification.

\begin{footnotesize}
\alglanguage{pascal}
\begin{algorithm}
\caption{\method{} training procedure}
\begin{algorithmic}[1]
\State $f_{\vomega_{\textrm{MAP}}}$, : pre-trained  network ,$\lambda$ : learning rate, nb\_epoch : number of epoch
\State \purple{\textbf{(Step 2:  adapt the DNN to a DNN)}} 
\State \small{\gray{\emph{\# Build a list of all the normalisation layers}}}
\State normalisation=[Batch\_normalisation,
Layer\_normalisation,
Instance\_normalisation]
\For{ layer \in\mathrm{f_{\vomega_{\textrm{MAP}}}.layers:}}{} 
\Begin
    \State \purple{\textbf{(Transform all Normalization Layers)}} 
    \If{(\mathrm{layer} \in normalisation):}
        \State replace layer by BNL
    \EndIf
\End
\State \purple{\textbf{(Step 3:  train \method{})}} 
\State t=0
\For{(epoch) \in \mathrm{nb\_epoch:}}{}
\Begin
    \For{(x,y) \in \mathrm{trainloader:}}{}
    \Begin
        \State \purple{\textbf{(Forward pass)}} 
        \State $\forall x_i \in B(t)$ calculate $f_{\vomega(t)}(x_i)$ 
        \State evaluate the loss $\mathcal{L}(\vomega(t),B(t)) $ 
        \State $\vomega(t) \leftarrow \vomega(t-1)-\lambda \nabla \mathcal{L}_{\vomega(t)}$
        \State \purple{\textbf{(Step update)}}
        \State  $t \leftarrow t+1$
    \End
\End
\end{algorithmic}
\label{algopseudo}
\end{algorithm}
\end{footnotesize}

\subsection{Theoretical analysis}  

Our approach raises several theoretical questions. In the supplementary material - as detailed in Section \ref{sup:Stability} - we show that \method{} exhibits greater stability than classical BNNs. Indeed, in variational inference BNNs, the gradients vary greatly: $\vepsilon$, crucial for obtaining the Bayesian interpretation, often introduces instability, perturbating the training. \method{} reduces this burden by applying this term on the latent space rather than the weights, thereby reducing the variance of the gradients, as empirically demonstrated in Appendix \ref{sup:Stability}.

Another question concerns the theoretical need to modify the loss and add the second term $\mathcal{E}$. We show in Appendix~\ref{sup:Multimodes} that it is theoretically sound in the case of a convex problem. Given that DNN optimization is inherently non-convex, adding this term may be theoretically debatable. However, a sensitivity analysis of this term - developed in Appendix~\ref{sec:ablation} - shows empirical benefits for performance and uncertainty quantification 

Finally, we discuss the challenge of estimating the equivalent BNNs to our networks in Appendix \ref{sup:Bayesiandiscuss}. Despite the theoretical value this information could provide concerning the posterior, it remains unused in practice. We solely sample the $\vepsilon$ and average over multiple training terms to generate robust predictions during inference.

%% file: sec/3_Experiments.tex
\section{Experiments \& Results\label{sec:xp_results}}

We test \method{} on image classification and semantic segmentation tasks. In each case, we measure metrics relative to the performance of the models but measure their uncertainty quantification abilities. All our models are implemented in PyTorch and trained on a single Nvidia RTX 3090. Appendix~\ref{sec:Implementation} details the hyper-parameters used in our experiments across architectures and datasets.

\subsection{Image classification}

\input{sec/tab_ic_cifar}

\noindent\textbf{Datasets.} We demonstrate the efficiency of \method{} on different datasets and backbones. We start with CIFAR-10 and CIFAR-100~\cite{krizhevsky2009learning} with ResNet-50~\cite{he2016deep} and WideResNet28-10~\cite{zagoruyko2016wide}. We then report results for \method{} on ImageNet~\cite{deng2009imagenet} with ResNet-50 and ViT~\cite{dosovitskiy2020image}. In the former case, we train all models from scratch. In the latter, we start from torchvision pre-trained models~\cite{paszke2019pytorch}.

\noindent\textbf{Baselines.} We compare \method{} against Deep Ensembles~\cite{lakshminarayanan2017simple} and four other ensembles: BatchEnsemble~\cite{wen2019batchensemble}, MIMO~\cite{havasi2021training}, Masksembles~\cite{durasov2021masksembles}, and  Laplace~\cite{laplace2021}.

\noindent\textbf{Metrics.} We evaluate the performance on classification tasks with the accuracy (Acc) and the Negative Log-Likelihood (NLL). We complete these metrics with the expected top-label calibration error (ECE)~\cite{naeini2015obtaining} and measure the quality of the OOD detection using the Areas Under the Precision/Recall curve (AUPR) and the operating Curve (AUC), as well as the False Positive Rate at $95\%$ recall (FPR95) similarly to Hendrycks et al.~\cite{hendrycks2016baseline}. We express all metrics in \%.

\noindent\textbf{OOD detection datasets.} For OOD detection tasks on CIFAR-10 and CIFAR-100, we use the SVHN dataset~\cite{netzer2011reading} as the out-of-distribution dataset and transform the initial problem into binary classification between in-distribution and out-of-distribution data using the maximum softmax probability as the criterion. For ImageNet, we use Describable Texture~\cite{wang2022vim} as the out-of-distribution dataset.

\noindent\textbf{Results.}
Tables \ref{tab:classif} and \ref{tab:classifINET} present the performance of \method{} across various architectures for CIFAR-10/100 and ImageNet, respectively.
Notably, \method{} consistently surpasses Laplace approximation and single-model baselines on most datasets and architectures. Furthermore, \method{} exhibits competitive performance compared to Deep Ensembles, achieving equivalent results with a similar number of parameters and training time to a single model. These findings underscore \method{} as a powerful and efficient method, demonstrating superior uncertainty quantification capabilities in image classification tasks while being easier to train.

\subsection{Semantic segmentation}

For the semantic segmentation part, we compare \method{} 
 against MCP~\cite{hendrycks2016baseline}, Deep Ensembles ~\cite{lakshminarayanan2017simple}, MC Dropout \cite{gal2016dropout}, TRADI~\cite{franchi2020tradi},  MIMO~\cite{havasi2021training} and LP-BNN~\cite{franchi2020encoding}, on StreetHazards~\cite{hendrycks2019anomalyseg},  BDD-Anomaly~\cite{hendrycks2019anomalyseg}, and MUAD~\cite{franchi2022muad} that allows comparison on diverse uncertainty quantification aspects of semantic segmentation.

\noindent\textbf{StreetHazards~\cite{hendrycks2019anomalyseg}.} StreetHazards is a large-scale synthetic dataset comprising various images depicting street scenes. The dataset consists of $5,125$ images for training and an additional $1,500$ images for testing. The training dataset has pixel-wise annotations available for 13 different classes. The test dataset is designed with 13 classes seen during training and an additional 250 out-of-distribution (OOD) classes that were not part of the training set. This diverse composition allows for assessing the algorithm's robustness in the face of various potential scenarios. In our experiments, we employed DeepLabv3+ with a ResNet-50 encoder, as introduced by Chen et al.~\cite{chen2018encoder}.

\noindent\textbf{BDD-Anomaly~\cite{hendrycks2019anomalyseg}.} BDD-Anomaly, a subset of the BDD100K dataset~\cite{yu2020bdd100k}, comprises $6,688$ street scenes for training and an additional $361$ for the test set. Within the training set, pixel-level annotations are available for $17$ distinct classes. The test dataset consists of the same $17$ classes seen during training and introduces 2 out-of-distribution (OOD) classes: motorcycle and train. In our experimental setup, we adopted DeepLabv3+\cite{chen2018encoder} and followed the experimental protocol outlined in\cite{hendrycks2019anomalyseg}. Similar to previous experiments, we utilized a ResNet-50 encoder~\cite{he2016deep} for the neural network architecture.

\begin{table}[t]
    \centering
    \resizebox{0.48\textwidth}{!}
    {
        \begin{tabular}{@{}llccccc<{\kern-\tabcolsep}}
            \toprule
            & {\textbf{Method}}  & \textbf{Acc $\uparrow$}  & \textbf{ECE $\downarrow$}  & \textbf{AUPR $\uparrow$} & \textbf{AUC $\uparrow$} & \textbf{FPR95 $\downarrow$}  \\
            
            \midrule
            
            \multirow{6}{*}{\rotatebox[origin=c]{90}{\textbf{ResNet-50}}} &
            Single Model & 77.8 & 12.1 & 18.0 & 80.9 & 68.6  \\
            
            & BatchEnsemble & 75.9 & \first \textbf{3.5} & \first \textbf{20.2} & 81.6 & 66.5  \\
            
            & MIMO ($\rho=1$) & 77.6 & 14.7 & 18.4 & 81.6 & 66.8  \\

            & Deep Ensembles & 79.2 & 23.3 & \second 19.6 & \first \textbf{83.4} & \first \textbf{62.1}  \\
            
            & Laplace & \first  \textbf{80.4} & 44.3 & 13.9 & 75.9 & 82.8 \\
            
            & {\method{}} & \second 79.5 & \second 9.65 & 17.8 & \second 82.0 & \second 65.2  \\
            
            \midrule
            
            \multirow{4}{*}{\rotatebox[origin=c]{90}{\textbf{ViT}}} &
            Single Model & 80.0 & \second 5.2 & 19.5 & 84.1 & \second 58.5   \\
            
            & {Deep Ensembles} & \first \textbf{81.7} & 13.5 & \second 21.7 & \first \textbf{85.5} & 60.3 \\

            & Laplace & \second 81.0 & 10.8 & \first \textbf{22.1} & 83.1 & 70.6 \\

            & {\method{}} & 80.6 & \first \textbf{4.32} & \second 21.7 & \second 85.4 & \first \textbf{55.1}  \\    
            
            \bottomrule
        \end{tabular}
    }
    \caption{\textbf{Performance on ImageNet using ResNet-50 and ViT} concerning in distribution and out-of-distribution metrics.}  \label{tab:classifINET}
\end{table}

\noindent\textbf{MUAD~\cite{franchi2022muad}.} MUAD consists of 3,420 images in the training set and 492 in the validation set. The test set comprises 6,501 images, distributed across various subsets: 551 in the normal set, 102 in the normal set with no shadow, 1,668 in the out-of-distribution (OOD) set. All these sets cover both day and night conditions, with a distribution of 2/3 day images and 1/3 night images. MUAD encompasses 21 classes, with the initial 19 classes mirroring those found in CityScapes~\cite{cordts2016cityscapes}. Additionally, three classes are introduced to represent object anomalies and animals, adding diversity to the dataset.
In our first experiment, we employed a DeepLabV3+ \cite{chen2018encoder} network with a ResNet50 encoder\cite{he2016deep} for training on MUAD.

\noindent\textbf{Results.} 
Table \ref{table:outofditribution} presents the results of \method{}, compared to various baselines on the three datasets. \method{} performs competitively with Deep Ensembles, a technique known for accurately quantifying uncertainty. Moreover, our approach exhibits faster training times, making it potentially more appealing for practitioners. We have not included a comparison with Laplace Approximation, as it is not commonly applied to semantic segmentation, and adapting DNNs for Laplace Approximation is not straightforward.

\begin{table}[t]
    \renewcommand{\figurename}{Table}
    \renewcommand{\captionfont}{\small}
    \begin{center}
        \resizebox{0.48\textwidth}{!}
        {
            \begin{tabular}{@{}clccccc<{\kern-\tabcolsep}}
                \toprule
                
                 & \textbf{Method}  & \textbf{mIoU} $\uparrow$ & \textbf{AUPR} $\uparrow$ & \textbf{AUC} $\uparrow$   & \textbf{FPR95} $\downarrow$ & \textbf{ECE} $\downarrow$ \\ 
                
                \midrule
                
                \multirow{7}{*}{\rotatebox[origin=c]{90}{\textbf{StreetHazards}}}  & Single Model & 53.90  & 6.91 & 86.60  & 35.74 & 6.52 \\ 
                
                & TRADI &  52.46	& 6.93 & 87.39  & 38.26 &6.33\\
                
                & Deep Ensembles& \second 55.59 & \first \textbf{8.32} & 87.94  & \first \textbf{30.29} & \second 5.33 \\
                
                & MIMO & 55.44 &  6.90 &  87.38  & 32.66 & 5.57 \\ 
                
                & BatchEnsemble  & \first \textbf{56.16}  & 7.59 & 88.17  &32.85 & 6.09 \\
                
                & LP-BNN &  54.50 & 7.18 & \second 88.33  &32.61 & \first \textbf{5.20}  \\
                
                &\method{} (ours)  & 53.82 & \second 7.85 &  \first \textbf{88.39}     & \second 32.02  & 6.09 \\ 
                
                \midrule 
                
                \multirow{7}{*}{\rotatebox[origin=c]{90}{\textbf{BDD-Anomaly}}}                               & Single Model &  47.63  & 4.50 & 85.15    & \second 28.78  & 17.68 \\ 
                
                & TRADI   & 44.26	& 4.54	& 84.80	& 36.87	& 16.61 \\ 
                
                & Deep Ensembles& \first \textbf{51.07}  & \second 5.24 & 84.80    & \first \textbf{28.55 } &\second 14.19 \\
                
                & MIMO& 47.20  & 4.32 &  84.38    & 35.24 &16.33  \\
                
                & BatchEnsemble  &  48.09  & 4.49 & 84.27    & 30.17  & 16.90  \\ 
                
                & LP-BNN  & \second 49.01  & 4.52 & \second 85.32 & 29.47  & 17.16  \\ 
                
                &\method{} (ours)  & 48.76 & \first \textbf{5.98} & \first \textbf{85.74} & 29.01  & \first \textbf{14.03} \\
                
                \midrule
                
                \multirow{6}{*}{\rotatebox[origin=c]{90}{\textbf{MUAD}}}  &  Single Model &  57.32 & \second 26.04 & 86.24 & 39.43 & 6.07 \\

                & MC-Dropout &  55.62 & 22.25 & 84.39 & 45.75 & 6.45 \\
                 & Deep Ensembles &  \second 58.29 & \first \textbf{28.02} & \second 87.10 &  37.60 & 5.88 \\
                & BatchEnsemble  &  57.10 & 25.70 & 86.90 &  38.81 &  6.01 \\
                & MIMO  &  \second 57.10 & 24.18 & 86.62 &  \second 34.80 & \second 5.81 \\

                &\method{} (ours)  & \first \textbf{61.96} & 24.37 & \first \textbf{91.55} & \first \textbf{21.68} & \first \textbf{5.58} \\
                
                \bottomrule
            \end{tabular}
        } %
     \end{center}
     \caption{{\textbf{Comparative results on the OOD task for semantic segmentation.}  We run all methods in similar settings using publicly available code for related methods. Results are averaged over three seeds. The architecture is a DeepLabv3+ based on ResNet50.}
     \label{table:outofditribution}}
     \vspace{-2mm}
 \end{table}

\section{Discussions\label{sec:Discussions}}

\subsection{General discussions}
We develop several discussions in the supplementary materials. First, we explore the theoretical aspects, including the stability of the DNNs in Appendix~\ref{sup:Stability}, the importance of multi-mode in Section~\ref{sup:Multimodes}, and the relationship with classical BNNs in Appendix~\ref{sup:Bayesiandiscuss}.
We experiment on the transfer of ViT-B-16 from Imagenet 21k~\cite{ridnik2021imagenet21k} to CIFAR-100 which highlights the potential of \method{} in transfer learning, achieving an accuracy of 92.18\%.
Additionally, we perform several ablation studies, notably on the impact of discarding multi-mode or the loss term $\mathcal{E}$ (defined in Section \ref{sec:loss}) in Appendix~\ref{sec:ablation}. We show that discarding $\mathcal{E}$ reduces the performance while incorporating the multi-mode improves uncertainty quantification.
Moreover, in Section~\ref{sec:grad_exp}, we analyze the variance of the gradients, confirming that our technique exhibits lower gradients than BNNs, making it more stable and easier to train.
Finally, Appendix~\ref{sec:stability_ABNN} delves into the variability of our method under different scenarios, exploring cases where we initiate \method{} from a single model and optimize from various initial checkpoints. Although our technique inherits the instabilities of the DNN, we observe that the standard variation is five times larger than that of the single model, indicating less stability than a standard DNN.

\subsection{Diversity of \method{}}

\begin{figure}[t!]
    \begin{subfigure}[b]{0.48\textwidth}
         \centering
         \includegraphics[width=\textwidth]{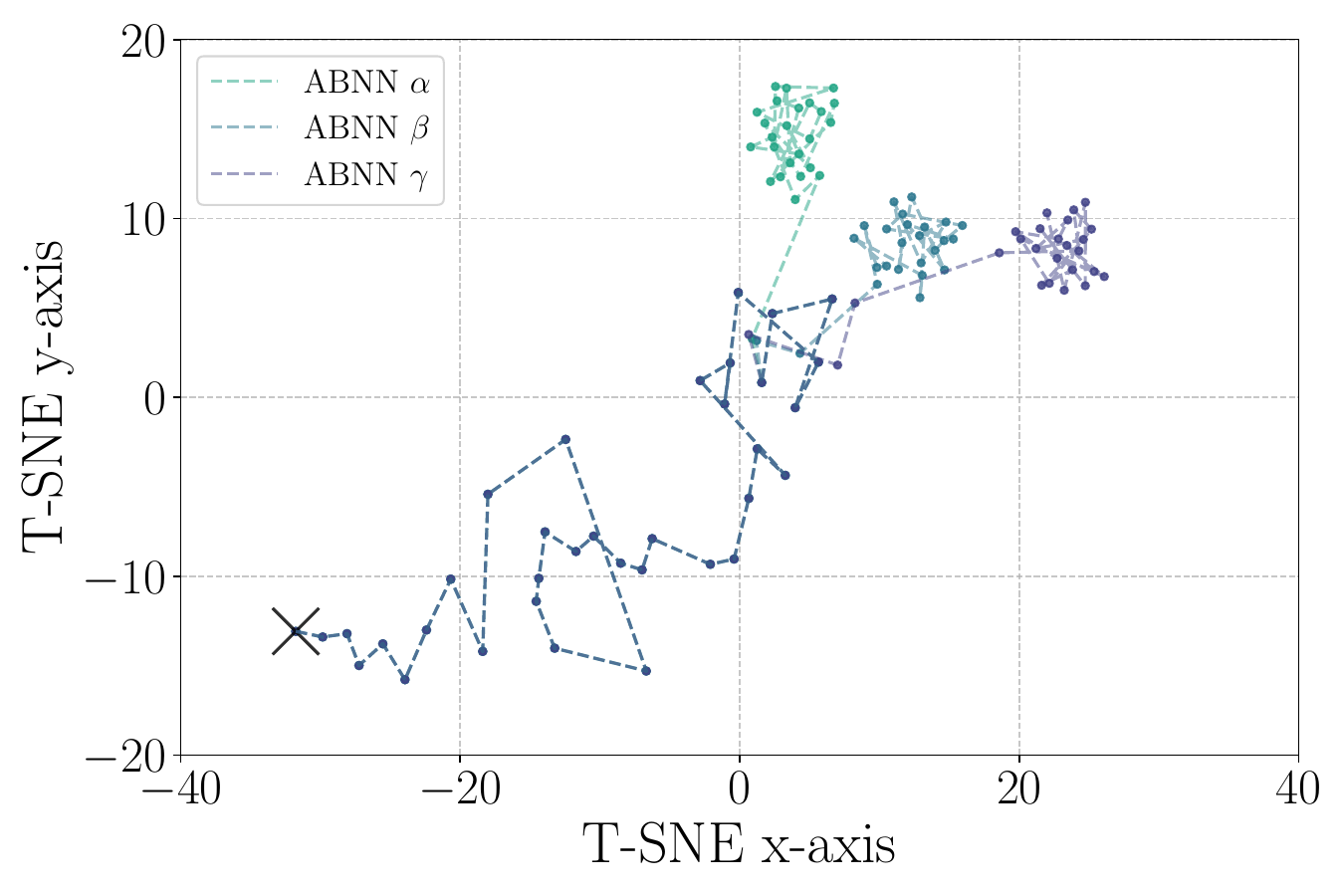}
         \caption{\method{}}
     \end{subfigure}
     \hfill
     \begin{subfigure}[b]{0.48\textwidth}
         \centering
         \includegraphics[width=\textwidth]{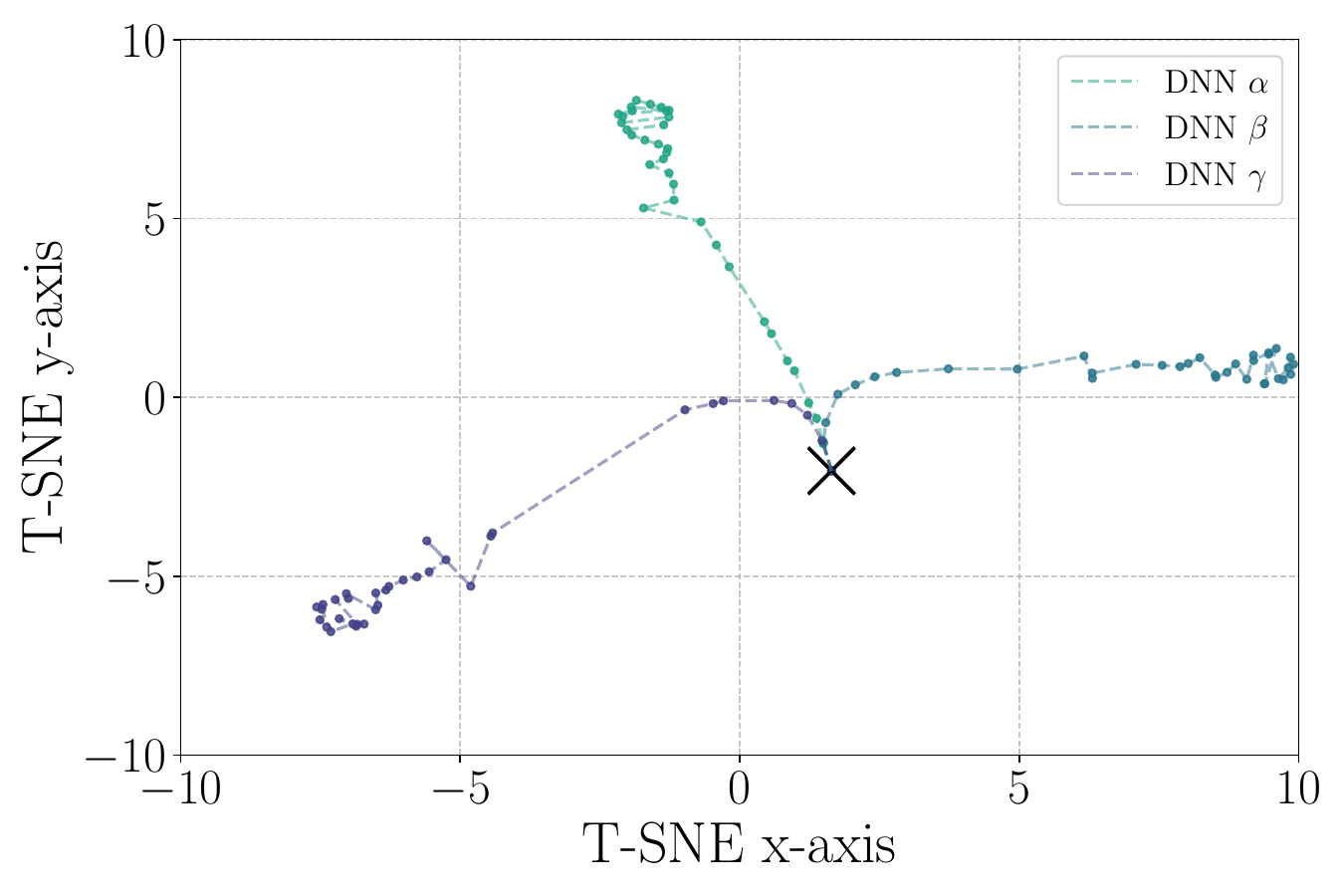}
         \caption{Deep Ensembles}
     \end{subfigure}
    \caption{\textbf{Comparison of the diversities of \method{} and Deep Ensembles~\cite{lakshminarayanan2017simple}}. T-SNE plot of the 20 principal components of the logits generated from 384 images for \method{} (a) and Deep Ensembles (b).}
    \label{fig:diversity_tsne}
\end{figure}

Concerning the diversity, we train \method{} on CIFAR-10 using a ResNet-50 architecture %
Specifically, we optimize a ResNet for 200 epochs and then fine-tune three ABNNs, starting from the optimal checkpoint. Additionally, we train two other ResNet-50s on CIFAR-10 to form a proper Deep Ensembles. As depicted in Figure~\ref{fig:diversity_tsne}, ABNN does not exhibit the same level of diversity as the Deep Ensembles. However, it is intriguing that even when initiated from a single DNN, \method{} manages to depart from its local minimum and explore different modes. This concept of different modes is further supported by Section~\ref{sec:sup:diversity}, where we analyze the mutual information of various ABNN checkpoints.

\section{Conclusion}

In conclusion, our proposed approach, \method{}, introduces a novel perspective to uncertainty quantification. Leveraging the strengths of pre-trained deterministic models, \method{} strategically transforms them into Bayesian %
networks with minimal modifications, offering enhanced stability and efficient posterior exploration. Through comprehensive experimental analyses, we demonstrate the effectiveness of \method{} both in predictive performance and uncertainty quantification, showcasing its potential applications in diverse scenarios.

The multi-mode characteristic of \method{}, coupled with a carefully designed loss function, not only addresses the challenge of the multi-modality of the posterior but also provides a stable and diverse ensemble of models. Our empirical evaluations on various datasets and architectures highlight the superiority of \method{} over traditional BNNs and post-hoc uncertainty quantification methods such as the Laplace approximation and showcase its competitiveness %
compared to state-of-the-art such as Deep Ensembles.

Moreover, \method{} exhibits promising results in transfer learning scenarios, underscoring its versatility and potential for broader applications. The insights gained from theoretical discussions and ablation studies further elucidate the underlying mechanisms of \method{}, contributing to a deeper understanding of its behavior and performance.

In summary, \method{} emerges as a robust and flexible solution for uncertainty-aware deep learning, offering a pragmatic bridge between deterministic and Bayesian paradigms. Its simplicity in implementation, coupled with superior performance and stability, positions \method{} as a valuable tool in the contemporary landscape of machine learning and Bayesian modeling.

\section*{Acknowledgments}

This work was performed using HPC resources from GENCI-IDRIS (Grant 2022 - AD011011970R2) and (Grant 2023 - Grant 2022 - AD011011970R3).

%% file: sec/tab_ic_cifar.tex
\begin{table*}[t]
    \centering
    \resizebox{0.74\textwidth}{!}
    {
    \begin{tabular}{@{}lllcccccccc<{\kern-\tabcolsep}}
    \toprule
        & & {\textbf{Method}} & \textbf{Acc} $\uparrow$ & \textbf{NLL} $\downarrow$ & \textbf{ECE} $\downarrow$ & \textbf{AUPR} $\uparrow$ & \textbf{AUC} $\uparrow$ & \textbf{FPR95} $\downarrow$ & $\mathbf{\Delta}$\textbf{Param} $\downarrow$ & \textbf{Time} (h) $\downarrow$ \\
        \midrule
        \multirow{14}{*}{\rotatebox[origin=c]{90}{\textbf{CIFAR-10}}} & \multirow{7}{*}{\rotatebox[origin=c]{90}{\textbf{ResNet-50}}} & {Single Model}   & 95.1 & 0.211 & 3.1 & 95.2 & 91.9 & 23.6 & $\varnothing$ & \first \textbf{1.7}\\
        & & {BatchEnsemble}   & 93.9 & 0.255 & 3.3 & 94.7 & 91.3 & 20.1 & \second 0.11 & 17.2 \\
        & & {MIMO ($\rho=1$)}   & \second 95.4 & 0.197 & 3.0 & 95.1 & 90.8 & 26.0 & \first \textbf{0.07} & 6.7 \\
        & & LPBNN & 95.0 & 0.251 & 9.4 & \first \textbf{98.4} & \first \textbf{96.9} & \first \textbf{10.3} & 1.83 & 17.2 \\
        & & {Deep Ensembles}   & \first  \textbf{96.0} &  \first \textbf{0.136} &  \first \textbf{0.8} & \second 97.0 & \second 94.7 & 15.5 & 70.56 & 6.8  \\
        & & Laplace & 95.3 & \second 0.160 & 1.3 & 96.0 & 93.3 & 18.8 & / & \first \textbf{1.7} \\
        & & {\method{}}   & \second 95.4 & 0.215 & \second 0.845 & \second 97.0 & \second 94.7 & \second 15.1 & 0.16 & \second 2.0 \\   
        
        \cmidrule[0.5pt](l){2-11}
        
        & \multirow{7}{*}{\rotatebox[origin=c]{90}{\textbf{WideResNet-28$\times$10}}} & {Single Model}  & 95.4 & 0.200 & 2.9 & 96.1 & 93.2 & 20.4 & $\varnothing$ & \first \textbf{4.2} \\
        & & {BatchEnsemble}  & \second 95.6  & 0.206 & 2.7  & 95.5 & 92.5 & 22.1 & \second 0.10 & 25.6 \\
        & & {MIMO ($\rho=1$)}  & 94.7  & 0.234 & 3.4 & 94.9 & 90.6 & 30.9 & 0.12 & 12.6 \\
        & & LPBNN & 95.1 & 0.249 & 2.9 & 95.4& 91.2& 29.5 & 0.71 & 23.3 \\
        & & {Deep Ensembles}  & \first \textbf{95.8} & \first \textbf{0.143} & \second 1.3 & \second 97.8 & \second 96.0 & \first \textbf{12.5} & 109.47 & 16.6 \\
        & & Laplace & \second 95.6 & \second 0.151 & \first \textbf{0.8} & 95.0 & 90.7 & 31.9 & / & \first \textbf{4.2} \\
        & & {\method{}}  & 93.7 & 0.198 &  1.8 &  \first\textbf{98.5} & \first\textbf{96.9} & \second 12.6 &  \first 0.05 & \second 5.0 \\
        
        \midrule
        
        \multirow{14}{*}{\rotatebox[origin=c]{90}{\textbf{CIFAR-100}}} & \multirow{7}{*}{\rotatebox[origin=c]{90}{\textbf{ResNet-50}}} &  {Single Model}   & 78.3 & 0.905 & 8.9 & 87.4 & 77.9 & 57.6 & $\varnothing$ & \first \textbf{1.7} \\

        & & {BatchEnsemble}    & 66.6  & 1.788 & 18.2 & 85.2 & 74.6  & 60.6 & \first 0.11 & 17.2 \\
        & & {MIMO ($\rho=1$)}    & \second 79.0  & \second 0.876 & 7.9 & 87.5 & 76.9 & 64.7 & 0.63 & 6.7 \\
        & & LPBNN & 78.5 & 1.02 & 11.3 & 88.2& 77.8& 73.5 & 1.83 & 17.2 \\
        & & {Deep Ensembles}   & \first \textbf{80.9} & \first \textbf{0.713} & \first \textbf{2.6} & \second 89.2 &  \second 80.8 & 52.5 & 71.12 & 6.8 \\
        & & Laplace & 78.2 & 0.987 & 14.2 & \second 89.2 & \first \textbf{81.0} & \second 51.8 & / & \first  \textbf{1.7}\\
        & & {\method{}}   & 78.9 & 0.889 & \second 5.5 & \first \textbf{89.4} & \first \textbf{81.0} & \first\textbf{50.1} & \second 0.16 & \second 2.0 \\
        
        \cmidrule[0.5pt](l){2-11}
        
        & \multirow{7}{*}{\rotatebox[origin=c]{90}{\textbf{WideResNet-28$\times$10}}} & {Single Model}  & 80.3 & 0.963 & 15.6 & 81.0 & 64.2 & 80.1 & $\varnothing$ & \first \textbf{4.2} \\
        & & {BatchEnsemble}   & \second 82.3  & 0.835 & 13.0  & \first\textbf{88.1} & \first\textbf{78.2} &  69.8 & \second 0.10 & 25.6 \\
        & & {MIMO ($\rho=1$)}  & 80.2  & \first \textbf{0.822} & \first\textbf{2.8} & 84.9 & 72.0 & 72.8 & 0.19 & 12.6 \\
        & & LPBNN & 79.7 & \second 0.831 & 7.0 & 79.0 & 70.1 & 71.4 & 0.72 & 23.3 \\ 
        & & {Deep Ensembles} & \first\textbf{82.5} & 0.903 & 22.9 & 81.6 & 67.9 & 71.3 & 109.64 & 16.6 \\
        & & Laplace & 80.1 & 0.942 & 16.0 & 83.4 & 72.1 & \second 59.9 & / & \first \textbf{4.2} \\
        & & {\method{}}  & 80.4 & 1.08 & \second 5.5 & \second 85.0 &  \second 75.0 & \first\textbf{57.7} & \first 0.05 & \second 5.0 \\
    \bottomrule
    \end{tabular}
    }
    \caption{
        \textbf{Performance comparison (averaged over five runs) on CIFAR-10/100 using ResNet-50 and Wide ResNet28$\times$10.} All ensembles have $M=4$ subnetworks. we highlight the best performances in bold. The ResNet-50 single model has respectively \{23.52, 23.70\}$\cdot10^6$ parameters. $\mathbf{\Delta}$Param is the number of parameters in excess of the corresponding method compared to the single model. Time is the training time in hours on a single RTX 3090.
    }
    \label{tab:classif}
\end{table*}

%% file: sec/X_suppl.tex
\startcontents

\twocolumn[{%
\renewcommand\twocolumn[1][]{#1}%
\maketitle
\maketitlesupplementary
\addtocontents{toc}{\protect\setcounter{tocdepth}{2}}
{
\hypersetup{linkcolor=black}
\printcontents{ }{1}{\section*{\contentsname}}{}
\vspace{8mm}
}}]

\clearpage  

The supplementary material encompasses 
multiple details and insights complementing the main paper as follows. In Section~\ref{sup:Notations}, we introduce and clarify the notations used throughout the 
paper. Theoretical insights are presented in Section~\ref{sup:Theoretical}, delving into the theoretical foundations of our methods. Section~\ref{sec:grad_exp} evaluates the stability of \method{}, shedding light on its robustness. Section~\ref{sec:stability_ABNN} shifts the focus to the stability of the training procedure, an essential aspect deserving exploration in every post-hoc technique. %
Section \ref{sec:ablation} delves into a sensitivity analysis and ablation study, exploring key components' resilience and performance 
impact. Section~\ref{sec:sup:diversity} focuses on the quality of the posterior estimated by \method{}. Finally, Sections \ref{sec:Implementation} and \ref{sup:Experiments} detail the training hyperparameters and showcase additional experiments, providing a comprehensive view of our methodology.

\section{Theoretical Analysis}\label{sup:Theoretical}

In this section, we develop a mathematical formalism to study the theoretical properties of \method{}. 
\subsection{Stability of \method{}}\label{sup:Stability}

Variational inference BNNs~\cite{blundell2015weight} are not commonly used in computer vision due to their challenges in scaling properly for deeper high capacity DNNs~\cite{dusenberry2020efficient}. In this section, we derive theoretical insights that entail the greater stability of \method{}, arguably also a Variational Inference BNN (VI-BNN). We start with deriving the gradients for a layer in a 
classic 2-hidden-layer MLP BNN. For the gradient of the loss on the mean of the weights of layer $j$, we have:

\small
\begin{multline}
\frac{\partial \mathcal{L}_{\scriptscriptstyle \text{MAP}}(\vomega) }{\partial W^{(j)}_{\mu,i,i'} }= \sum_{i''} \frac{\partial \mathcal{L}_{\scriptscriptstyle \text{MAP}}(\vomega) }{\partial \vh_{j+1,i''} } \\ \left[ W^{(j+1)}_{\mu,i'',i} +\mathcal{E}^{(j+1)}_{i'',i} W^{(j+1)}_{\sigma,i'',i} \right]  \frac{a'(\vh_{j,i})\va_{j-1,i'}}{\sigma_{j} }.
\end{multline}

On its standard deviation, we have:
\begin{multline}
\frac{\partial \mathcal{L}_{\scriptscriptstyle \text{MAP}}(\vomega) }{\partial W^{(j)}_{\sigma,i,i'} }= \sum_{i''} \frac{\partial \mathcal{L}_{\scriptscriptstyle \text{MAP}}(\vomega) }{\partial \vh_{j+1,i''} } \\ \left[ W^{(j+1)}_{\mu,i'',i} +\mathcal{E}^{(j+1)}_{i'',i} W^{(j+1)}_{\sigma,i'',i} \right]\frac{a'(\vh_{j,i})\vepsilon_{j,i,i'} \va_{j-1,i'}}{\sigma_{j} }.
\end{multline}
\normalsize

For the gradients of the \method{} parameters, in the case of a 2-hidden-layer MLP BNN, we have:
\begin{equation}
\frac{\partial \mathcal{L}_{\scriptscriptstyle \text{MAP}}(\vomega) }{\partial \gamma^{(j)}_{i} }= \sum_{i''} \frac{\partial \mathcal{L}_{\scriptscriptstyle \text{MAP}}(\vomega) }{\partial \vh_{j+1,i''} } W^{(j+1)}_{i'',i} \frac{\vh_j - \mu_{j}}{\sigma_{j} }(1+\vepsilon_{j,i})a'(\vu_{j,i}),
\end{equation}

as well as, on $\beta$,

\begin{equation}
\frac{\partial \mathcal{L}_{\scriptscriptstyle \text{MAP}}(\vomega) }{\partial \beta_{j,i} }= \sum_{i''} \frac{\partial \mathcal{L}_{\scriptscriptstyle \text{MAP}}(\vomega) }{\partial \vh_{j+1,i''} } W^{(j+1)}_{i'',i} a'(\vu_{j,i}).
\end{equation}
\normalsize

We have four random variables: $\frac{\partial \mathcal{L}_{\scriptscriptstyle \text{MAP}}(\vomega) }{\partial \vh_{j+1,i''} }$, $\mathcal{E}^{(j+1)}_{i'',i}$ and $\vepsilon_{j,i}$  along with $a'(\vu_{j,i})$. Let's consider calculating the conditional variance given $\frac{\partial \mathcal{L}{\scriptscriptstyle \text{MAP}}(\vomega) }{\partial \vh{j+1,i''} }$ for all $i''$. Assuming that all random variables associated with a single neuron are independent, we have:

\begin{multline}
\mbox{var}\left(\frac{\partial \mathcal{L}_{\scriptscriptstyle \text{MAP}}(\vomega) }{\partial W^{(j)}_{\mu,i,i'} }\right)= \sum_{i''} \Bigg(\frac{\partial \mathcal{L}_{\scriptscriptstyle \text{MAP}}(\vomega) }{\partial \vh_{j+1,i''} }  \frac{W^{(j+1)}_{\sigma,i'',i} \va_{j-1,i'}}{\sigma_{j} } \Bigg)^2\\ \times \mbox{var} \left[ \mathcal{E}^{(j+1)}_{i'',i}  a'(\vh_{j,i})\right] 
\end{multline}

and

\begin{multline}
\mbox{var}\left(\frac{\partial \mathcal{L}_{\scriptscriptstyle \text{MAP}}(\vomega) }{\partial W^{(j)}_{\sigma,i,i'} }\right)= \sum_{i''} \Bigg(\frac{\partial \mathcal{L}_{\scriptscriptstyle \text{MAP}}(\vomega) }{\partial \vh_{j+1,i''} } \\  \frac{W^{(j+1)}_{\sigma,i'',i} \va_{j-1,i'}}{\sigma_{j} } \Bigg)^2\mbox{var}\left[ \mathcal{E}^{(j+1)}_{i'',i}  a'(\vh_{j,i})\vepsilon_{j,i,i'}\right]. 
\end{multline}
Using the fact that $\vepsilon_{j,i,i'}$ is independent %
of $\mathcal{E}^{(j+1)}_{i'',i}$ and $  a'(\vh_{j,i})$ we have that 
\small
\begin{equation}
\mbox{var}\left[ \mathcal{E}^{(j+1)}_{i'',i}  a'(\vh_{j,i})\vepsilon_{j,i,i'}\right] {=} 
\mbox{var}\left[ \mathcal{E}^{(j+1)}_{i'',i}  a'(\vh_{j,i})\right] + \mathbb{E}\left(\mathcal{E}^{(j+1)}_{i'',i}  a'(\vh_{j,i})\right)^2 
\end{equation}

In the case of \method{}, we can express the conditional variance as follows:

\footnotesize
\begin{equation}
\mbox{var}\left(\frac{\partial \mathcal{L}_{\scriptscriptstyle \text{MAP}}(\vomega) }{\partial \gamma^{(j)}_{i} }\right){=} \sum_{i''} \left(\frac{\partial \mathcal{L}_{\scriptscriptstyle \text{MAP}}(\vomega) }{\partial \vh_{j+1,i''} } W^{(j+1)}_{i'',i} \frac{\vh_j - \mu_{\vh_j}}{\sigma_{j} } \right)^2\mbox{var}\left[\vepsilon_{j,i}a'(\vu_{j,i})\right]
\end{equation}

\normalsize
\begin{equation}
\mbox{var}\left(\frac{\partial \mathcal{L}_{\scriptscriptstyle \text{MAP}}(\vomega) }{\partial \beta_{j,i} }\right){=}\sum_{i''} \left(\frac{\partial \mathcal{L}_{\scriptscriptstyle \text{MAP}}(\vomega) }{\partial \vh_{j+1,i''} } W^{(j+1)}_{i'',i}\right)^2 \mbox{var}\left[a'(\vu_{j,i})\right]
\end{equation}

Assuming that $\mbox{var}\left[\vepsilon_{j,i}a'(\vu_{j,i})\right] = \mbox{var}\left[ \mathcal{E}^{(j+1)}_{i'',i}  a'(\vh_{j,i})\right] $, we find that the variances of $\mbox{var}\left(\frac{\partial \mathcal{L}_{\scriptscriptstyle \text{MAP}}(\vomega) }{\partial W^{(j)}_{\mu,i,i'} }\right)$  and $\mbox{var}\left(\frac{\partial \mathcal{L}_{\scriptscriptstyle \text{MAP}}(\vomega) }{\partial W^{(j)}_{\sigma,i,i'} }\right)$ are directly proportional to the variance of $\mbox{var}\left(\frac{\partial \mathcal{L}_{\scriptscriptstyle \text{MAP}}(\vomega) }{\partial \gamma_{j,i} }\right)$.This proportionality is associated with the magnitudes of the weight values, and we assume that they are of similar magnitude.
 Consequently, the variance of the gradient of the parameters $\beta_{j,i}$ is the smallest among all, followed by the variance of the parameters $\gamma_{j,i}$ and $W^{(j)}_{\mu,i,i'}$, which are roughly equivalent, with the highest variance observed for $W^{(j)}_{\sigma,i,i'}$. This property results in more stable backpropagation for \method{} compared to 
 classic VI-BNNs.

\subsection{Multi-modes with \method{}}\label{sup:Multimodes}

The posterior of the DNN often comprises multiple modes~\cite{wilson2020bayesian,laurent2023symmetry}, making it non-trivial for an unimodal distribution chosen to represent the BNN's posterior to account for these different modes effectively. One approach to address this issue is to train multiple BNNs, as proposed in the multi-SWAG method by Wilson and Izmailov~\cite{wilson2020bayesian}. However, adapting this strategy to VI-BNNs inherits the instability issue from 
classic BNNs and may struggle to fit 
multiple modes accurately.

Our solution, \method{}, also faces a similar challenge, where we need to ensure that the technique doesn't collapse into the same local minima during training. %
We introduce a small perturbation to the loss function to prevent this collapse, which helps diversify the 
optimization process. 
This perturbation involves modifying the class weights within the cross-entropy loss. 
More precisely, contrary to 
classic VI-BNN that optimizes the Evidence Lower Bound (ELBO) loss, we propose to maximize the MAP.
\methods{} optimize the following loss:
\begin{figure}[t!]
    \centering{\includegraphics[width=0.9\linewidth]{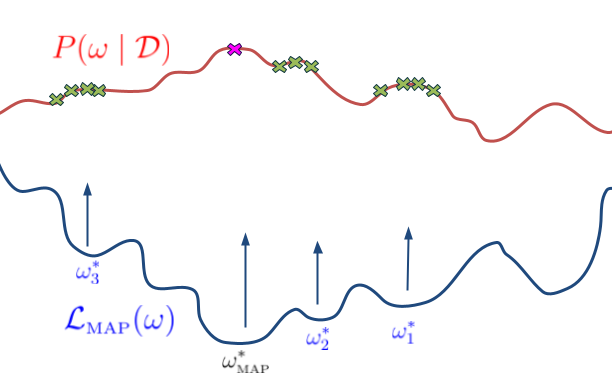}}
    \caption{Illustration of the training loss (in blue) and the corresponding posterior distribution (in red). Due to the multi-modal nature of the posterior, training multiple \methods{} with distinct final weights (such as $\vomega_1^*$, $\vomega_2^*$, and $\vomega_3^*$) enables sampling from different modes, enhancing the overall estimation of the posterior.}
    \label{fig:loss_posterior}
\end{figure}
\begin{equation}\label{eq:loss_MAP_multi}
\mathcal{L}(\vomega)
  = - \sum_{(\vx_i,y_i) \in \mathcal{D}} \alpha(y_i) \log P(y_i \mid \vx_i,\vomega) - \log P(\vomega),
\end{equation}
In the standard cross-entropy loss, all classes are given equal weight, typically represented as $\alpha(y_i) = 1$. However, our approach deliberately introduces the \emph{random prior}: random weight adjustments for certain classes, denoted as $\eta_i$ (such that $\alpha(y_i) = 1+\eta_i$). This manipulation encourages various \method{}s to specialize as experts in different classes. Consequently, the training loss is formulated as follows:
\begin{equation}\label{eq:loss_MAP_multi1_supp}
\mathcal{L}(\vomega)
  =\mathcal{L}_{\scriptscriptstyle \text{MAP}}(\vomega) + \mathcal{E}(\vomega)
\end{equation}
where 

\begin{equation}\label{eq:RP}
\mathcal{E}(\vomega) = -\sum_{(\vx_i,y_i) \in \mathcal{D}} \eta_i \log P(y_i \mid \vx_i,\vomega).
\end{equation}
 Let's denote $\vomega^{(0)}$ as the parameter configuration that minimizes $\mathcal{L_{\scriptscriptstyle \text{MAP}}}$.
Let us suppose for simplicity that the loss function is convex to provide a theoretical grounding to the random prior. After a single step of gradient descent (GD), we have:
\begin{equation}\label{eq:loss_MAP_multi2}
\vomega^{(1)} = \vomega^{(0)} -\lambda\nabla\mathcal{L}(\vomega^{(0)}),
\end{equation}
where $\vomega^{(1)}$ represents the parameters after the first optimization step, the superscript denotes the iteration number, and $\lambda$ is the learning rate.
 
We can express the updated loss $\mathcal{L}(\vomega^{(1)})$ using the GD, as shown in Eq.~\eqref{eq:loss_MAP_multi3}:
\begin{equation}\label{eq:loss_MAP_multi3}
\mathcal{L}(\vomega^{(1)}) = \mathcal{L}(\vomega^{(0)} -\lambda\nabla\mathcal{L}(\vomega^{(0)}))
\end{equation}
Now, by applying a first-order Taylor expansion to $\mathcal{L}$, we can express the updated loss $\mathcal{L}(\vomega^{(1)})$ as a function of the initial loss $\mathcal{L}(\vomega^{(0)})$ and the gradient update, as shown in Eq.~\eqref{eq:loss_MAP_multi4}:
\begin{equation}\label{eq:loss_MAP_multi4}
\mathcal{L}(\vomega^{(1)}) = \mathcal{L}(\vomega^{(0)}) -\lambda\nabla\mathcal{L}(\vomega^{(0)})^t\nabla\mathcal{L}(\vomega^{(0)})
\end{equation}
This equation can be further simplified 
by noticing that $\nabla\mathcal{L_{\scriptscriptstyle \text{MAP}}}(\vomega^{(0)})=0$:
\begin{equation}\label{eq:loss_MAP_multi5}
\mathcal{L}(\vomega^{(1)}) = \mathcal{L}(\vomega^{(0)}) -\lambda\nabla\mathcal{E}(\vomega^{(0)})^t\nabla\mathcal{E}(\vomega^{(0)})
\end{equation}

Starting with Eq.~\eqref{eq:loss_MAP_multi5}, we have:
\begin{multline}\label{eq:loss_MAP_multi6}
    \mathcal{L}(\vomega^{(1)}) = \mathcal{L}(\vomega^{(0)}) - \lambda \sum_{(\vx_i,y_i)  \in \mathcal{D}} \sum_{(\vx_{i'},y_{i'})  \in \mathcal{D}}\eta_{i}\eta_{i'} \\  \log P(y_{i'} \mid \vx_{i'},\vomega^{(0)})  \log P(y_i \mid \vx_i,\vomega^{(0)})
\end{multline}
    
Under the assumption that $\lambda\eta_{i}\eta_{i'}$ is small and that $\log P(y_{i'} \mid \vx_{i'},\vomega)$ is bounded, we can approximate the loss as follows: $\mathcal{L}(\vomega^{(1)}) \simeq \mathcal{L}(\vomega^{(0)})$. Consequently, we have $\nabla\mathcal{L}(\vomega^{(1)}) \simeq \nabla\mathcal{L}(\vomega^{(0)})$, and after another optimization step, $\vomega(2)$ is updated as $\vomega(2) = \vomega^{(1)} - \lambda\nabla\mathcal{L}(\vomega^{(1)})= \vomega^{(0)}  - 2\lambda\nabla\mathcal{L}(\vomega^{(0)})$.

By applying the same technique iteratively, for all $t$, we can approximate the loss as $\mathcal{L}(\vomega^{(t)}) \simeq \mathcal{L}(\vomega^{(0)})$. This leads also to the relationship $\vomega^{(t)} = \vomega^{(0)} - t\lambda\nabla\mathcal{L}(\vomega^{(0)})$.

Under the conditions that the loss is convex, the derivatives of the DNN are bounded, and $\lambda\eta_{i}\eta_{i'}$ is small, we can find minima of $\mathcal{L}$ with similar loss values by introducing weight diversity. This loss function can be valuable in encouraging each DNN to escape from the global minima, particularly in convex cases. In non-convex cases, standard SGD may already help escape from local minima, but this additional loss may offer extra assistance in avoiding the same local minima.
In Figure \ref{fig:loss_posterior}, we present a visualization depicting the training loss alongside the posterior distribution. This Figure highlights the importance of training multiple \method{}s with different optimal solutions to improve the quality in estimating the posterior distribution.

\subsection{Discussion on the Bayesian Neural Network Nature of \method{}}\label{sup:Bayesiandiscuss}

The Law of the Unconscious Statistician (LOTUS)~\cite{feller1991introduction} is a theorem in probability theory that offers a method for computing the expected value of a function of a random variable. Hence, for a continuous random variable $X$ with a probability density function $f_X(x)$, the expected value of a function $Y=g(X)$ is expressed as:

\begin{equation}\label{eq:LOTUS1}
E_Y(Y)=E_X[g(X)]
\end{equation}

In our scenario, let $U_j$ denote the random variable associated with $\vu_j$, and $W^{(j)}$ represent the random variable $W^{(j)}$ (we use the same letter for simplification). Thus, we have:

\begin{equation}\label{eq:LOTUS2}
E_{U_j}(U_j)= \frac{\vh_j - \mu_{\vh_j}}{\sqrt{\sigma_{\vh_j}^2 + \epsilon} }\gamma_j+\beta_j
\end{equation}

For simplification, we set $\beta_j$=0, which leads to

\begin{equation}\label{eq:LOTUS3}
\mbox{var}_{U_j}(U_j)=\left( \frac{\vh_j - \mu_{\vh_j}}{\sqrt{\sigma_{\vh_j}^2 + \epsilon} }\gamma_j \right)^2
\end{equation}

Here, the parameters $\gamma_j$ and $\beta_j$ are optimized to obtain the best Bayesian Neural Network (BNN)

\section{Experiment on the variance of the gradient}\label{sec:grad_exp}

To validate our hypothesis that \method{} is more stable than VI-BNNS, as detailed in section \ref{sup:Stability}, we analyze the variances of the gradients of 
classic DNNs, VI-BNNs, and \methods{}. Table~\ref{table:gradSTD} reveals that the gradient variance of BNNs are significantly greater than that of DNNs, aligning with the inherent challenges in training BNNs. Notably, \method{} exhibits a considerably lower gradient, stemming from only the weights of BNL are trained. Both empirical observations and theoretical considerations affirm the superior stability of \method{} on this side. Additionally, Table~\ref{table:gradSTD} includes results for a VI-BNN, which, as discussed in this section, exhibits suboptimal performance. Noteworthy 
other works~\cite{dusenberry2020efficient, hron2022wide} also express concerns regarding the stability of VI-BNNs.

\begin{table}[h!]
\renewcommand{\figurename}{Table}
\renewcommand{\captionfont}{\small}
\centering
    \resizebox{0.7\columnwidth}{!}
    {
    \begin{tabular}{@{}lc@{}}
        \toprule
        \textbf{Method} & \textbf{variance} ($10^{-4}$)  \\
        \midrule
        ResNet50 & 1.43  \\
        ResNet50 BNN & 2.54  \\
        ResNet50 ABNN & 3.20$\cdot10^{-2}$\\
        Wide-ResNet & 1.37  \\
        Wide-ResNet BNN & 2.53  \\
        Wide-ResNet ABNN & 9.91$\cdot10^{-2}$ \\
        \bottomrule
    \end{tabular}
    }
    \caption{
        \textbf{Variance of gradients on relevant parameters} \\ %
        Single model: every weight (no bias) \\
        BNN: means of the weight samplers \\
        ABNN: parameters linked to the BNL layers \\
    }
    \label{table:gradSTD}
\end{table}

\section{Discussion on stability  of the training of \method}\label{sec:stability_ABNN}

\method{} being a post-hoc technique, it is imperative to ensure that it does not introduce instability to DNNs, especially in the critical domain of uncertainty quantification. We train multiple single models based on a ResNet-50 architecture on CIFAR-10 to verify this point, calculating the standard deviation of the different metrics. Additionally, we derive several \methods{} starting from these checkpoints and assess the variance. Finally, we apply our technique to train an \method{} for each checkpoint of the single models. Table~\ref{table:stdTrain} demonstrates that our approach minimally increases the variance, 
confirming that it does not introduce instability to the uncertainty quantification process.

\begin{table}[t]
    \renewcommand{\figurename}{Table}
    \resizebox{\columnwidth}{!}
    {
    \centering
    \begin{tabular}{@{}lccccc@{}}
        \toprule
        & \textbf{Acc} & \textbf{ECE} & \textbf{AUPR} & \textbf{AUC} & \textbf{FPR95} \\
        \midrule
        Single model & 0.356 & 0.0002 & 1.036 & 1.445 & 2.740 \\
        one $\vomega_{\textrm{MAP}}$ + Multiple ABNN & 0.066 & 0.0009 & 0.130 & 0.224 & 0.658 \\
        Multiple  $\vomega_{\textrm{MAP}}$ + ABNN & 0.324 & 0.0011 & 1.131 & 1.570 & 3.202 \\
        \bottomrule
    \end{tabular}
    }
    \caption{
     \textbf{Standard Deviation (SD) Comparison of ABNN and DNN.} The first row presents the SD of a single DNN, while the second row depicts the SD of \method{} starting from a single checkpoint. The last row quantifies the SD of \method{} when trained from different checkpoints. All training scenarios use the optimal hyperparameters for \method{} on a ResNet-50 architecture on the CIFAR-10 dataset.
    }
    \label{table:stdTrain}
\end{table}

\section{Ablation study of \method{} and Sensitivity analysis}\label{sec:ablation}

In Table~\ref{tab:ablation}, we conduct a study to inspect the impact of adding or removing two characteristics from our method. First, we investigate whether the addition of the random prior (linked to $\mathcal{E}$ term in Eq. \eqref{eq:RP}), which introduces disturbance to the loss, improves the performance of \method{}. We test the Random Prior (RP) to check whether $\mathcal{E}$ contributes positively. Notably, when training a single RP model (when \textbf{MM} is \xmark), it appears to degrade performance as it corrupts the cross-entropy. Conversely, in the case of training multiple \methods{} (\textbf{MM} is \cmark), RP seems to improve uncertainty quantification metrics. The second aspect is the training of multiple modes: Table~\ref{tab:ablation} shows that incorporating multiple modes (\textbf{MM} is \cmark) 
improves the quality of the uncertainty quantification, in particular for OOD detection.

\begin{table*}[t]
    \centering
    \resizebox{0.6\textwidth}{!}
    {
    \begin{tabular}{@{}llccccccc<{\kern-\tabcolsep}}
    \toprule
        &   & \textbf{RP}  & \textbf{MM}  & \textbf{Acc} $\uparrow$ & \textbf{ECE} $\downarrow$ & \textbf{AUPR} $\uparrow$ & \textbf{AUC} $\uparrow$ & \textbf{FPR95} $\downarrow$  \\
        \midrule
        \multirow{8}{*}{\rotatebox[origin=c]{90}{\textbf{CIFAR-10}}} & \multirow{4}{*}{\rotatebox[origin=c]{90}{\small\textbf{ResNet-50}}} & \xmark & \xmark & 94.7 & 1.0 & 96.8 & 94.1 & 17.3 \\
        & & \cmark & \xmark & 95.2 & \second 0.89 & 96.7 & 94.4 & \second 15.3 \\
        & & \xmark & \cmark & \second 95.3 & 1.34 & \first \textbf{97.1} & \first \textbf{94.8} & 15.7 \\
         & & \cmark & \cmark & \first \textbf{95.4} & \first \textbf{0.845} & \second 97.0 & \second 94.7 & \first \textbf{15.1} \\
         
        \cmidrule[0.5pt](l){2-9}
        & \multirow{3}{*}{\rotatebox[origin=c]{90}{\small\textbf{WideResNet}}} & \xmark & \xmark & \second 94.4 & \first \textbf{1.34} & 96.4 & 93.7 & 18.2 \\
        & & \cmark & \xmark & 92.8 & 2.2 & \second 97.5 & \second 95.3 & \second 14.8 \\
        & & \xmark & \cmark & \first \textbf{94.9} & \second 1.38 & 97.3 & 95.1 & 15.7 \\
        & & \cmark & \cmark & 93.7 & 1.8 & \first \textbf{98.5} & \first \textbf{96.9} & \first \textbf{12.6} \\
        
        \cmidrule[0.5pt](l){1-9}
        \multirow{8}{*}{\rotatebox[origin=c]{90}{\textbf{CIFAR-100}}} & \multirow{4}{*}{\rotatebox[origin=c]{90}{\small\textbf{ResNet-50}}} & \xmark & \xmark & \first 78.8 & 5.7 & 89.3 & 80.8 & 50.4 \\
        & & \cmark & \xmark & 78.7& \first \textbf{5.5} & 89.4 & \second 81.0 & 50.1 \\
        & & \xmark & \cmark &  78.3 & 5.8 & \second 89.6 & \first \textbf{81.6} & \first \textbf{48.2} \\
        & & \cmark & \cmark & \first 78.8 & \second 5.6 & \first \textbf{89.7} & \first \textbf{81.6} & \second 49.0\\  

        \cmidrule[0.5pt](l){2-9}
        & \multirow{4}{*}{\rotatebox[origin=c]{90}{\small\textbf{WideResNet}}} & \xmark & \xmark & \first \textbf{80.5} & \second 5.6 & 84.0 & 72.6 & 62.2 \\
        & & \cmark & \xmark & 79.6 & \first \textbf{5.5} & 84.7 & 74.9 & \second 55.8 \\
        & & \xmark & \cmark & \second  80.4 & \first \textbf{5.5} & \second 85.0 & \second 75.0 & 57.7 \\
        & & \cmark & \cmark & 78.2 & 5.9 & \first \textbf{87.8} & \first \textbf{79.7} & \first \textbf{49.0} \\
               
        \bottomrule
    \end{tabular}
    }
    \caption{
        \textbf{Performance comparison (averaged over five runs) on CIFAR-10/100 using ResNet-50 and WideResNet28$\times$10.} \textbf{RP} is Random Prior, and \textbf{MM} is multi-mode. For OOD detection, we use the SVHN dataset.
    } \label{tab:ablation}
\end{table*}

\begin{table}[t]
    \resizebox{\columnwidth}{!}
    {
    \begin{tabular}{@{}lccccc<{\kern-\tabcolsep}}
        \toprule
        \textbf{Coef. LR} & \textbf{Acc} $\uparrow$ & \textbf{ECE} $\downarrow$ & \textbf{AUPR} $\uparrow$ & \textbf{AUC} $\uparrow$ & \textbf{FPR95} $\downarrow$ \\
        \midrule
        $\times10^{-1}$ & 95.5 & 0.9 & 96.2 & 93.0 & 20.7 \\
        $\times2\cdot10^{-1}$ & 95.4 & 0.9 & 96.5 & 93.8 & 18.2 \\
        $\times5\cdot10^{-1}$ & 95.4 & 1.0 & 96.3 & 93.3 & 19.9 \\
        $\times 1$ & 95.4 & 0.9 & 97.0 & 94.7 & 15.1 \\
        $\times 2$ & 95.3 & 1.1 & 95.7 & 92.3 & 22.0 \\
        $\times 5$ & 94.6 & 1.5 & 96.1 & 93.1 & 19.6 \\
        $\times 10$ & 93.9 & 1.1 & 96.7 & 94.2 & 19.2 \\
        \bottomrule
    \end{tabular}
    }
    \caption{
        \textbf{Sensitivity Analysis of the Learning Rate on CIFAR-10.} e conducted training for \method{} using a learning rate set at $0.0057$ multiplied by the \textbf{Coef LR}.
    }
    \label{tab:sentivity1}
\end{table}

\begin{table}[t]
    \resizebox{\columnwidth}{!}
    {
    \begin{tabular}{@{}lccccc<{\kern-\tabcolsep}}
        \toprule
        \textbf{Coef. LR} & \textbf{Acc} $\uparrow$  & \textbf{ECE} $\downarrow$ & \textbf{AUPR} $\uparrow$ & \textbf{AUC} $\uparrow$ & \textbf{FPR95} $\downarrow$ \\
        \midrule
        $\times10^{-1}$ & 79.0 & 5.4 & 89.0 & 80.3 & 51.5 \\
        $\times2\cdot10^{-1}$ & 78.9 & 5.5 & 88.6 & 79.9 & 52.5 \\
        $\times5\cdot10^{-1}$ & 78.9 & 5.6 & 88.9 & 80.0 & 52.5 \\
        $\times 1$ & 78.9 & 5.5 & 89.4 & 81.0 & 50.1 \\
        $\times 2$ & 78.8 & 5.7 & 88.6 & 79.8 & 52.6 \\
        $\times 15$ & 79.0 & 5.5 & 88.8 & 80.1 & 52.0 \\
        $\times 10$ & 78.8 & 5.7 & 88.8 & 80.2 & 51.4 \\
        \bottomrule
    \end{tabular}
    }
    \caption{
        \textbf{Sensitivity Analysis of the Learning Rate on CIFAR-100.} e conducted training for \method{} using a learning rate set at $0.00139$ multiplied by the \textbf{Coef. LR}.
    }
    \label{tab:sentivity2}
\end{table}

We analyze the performance variations in Tables~\ref{tab:sentivity1} and \ref{tab:sentivity2} by modifying the learning rate during the fine-tuning phase. It's essential to highlight that this hyperparameter is the only one of \method{}. We fine-tune a single model with various learning rates for this evaluation after adapting it to \method{}. Remarkably, the learning rate appears non-critical, as the performances on CIFAR-10 and CIFAR-100 exhibit minimal variations, around one percent.

\clearpage
\section{Discussion on \method{}'s diversity}\label{sec:sup:diversity}

We trained a Deep Ensembles of ResNet-50 architecture for 200 epochs, specifically three networks. As demonstrated in \cite{fort2019deep}, Deep Ensembles strikes a good balance between accuracy and diversity, enabling effective uncertainty quantification. Simultaneously, we train three \methods{}, initializing them from the same checkpoint. It's important to highlight that all \method{} parameters were trained for this experiment. To visually compare the results, akin to \cite{fort2019deep} (Figure 2.c), we conduct a t-SNE analysis on the latent space of all the checkpoints after each epoch. Initially, \method{} shows limited diversity due to having only one checkpoint. However, as training progresses, some diversity emerges, though less than observed in Deep Ensembles. Table \ref{table:MutualINFO} illustrates that \method{} exhibits 
lower mutual information than BNN on both in-distribution (IDs) and out-of-distribution (OOD) samples. Yet, interestingly, \method{} achieves a superior mutual information ratio on OODs/IDs. It's worth noting that mutual information serves as a metric for measuring diversity and can also quantify epistemic uncertainty. Mutual information is defined  for a finite set $\{\vomega_1, \ldots \vomega_M\}$ of $M$ weight configurations sampled from the posterior distribution as : 

\small

\begin{multline}\label{eq:MI1}
\underbrace{I\left( P(\vomega \mid \mathcal{D}) \right)}_{\mbox{Epistemic uncertainty}}   = \underbrace{\mathcal{H}( \frac{1}{M} \sum_{m=1}^M  P( y \mid \vx, \vomega_m))}_{\mbox{Total uncertainty}} \\ - \underbrace{\frac{1}{M} \sum_{m=1}^M  \mathcal{H}( P( y \mid \vx, \vomega_m))}_{\mbox{Aleatoric uncertainty}}.
\end{multline}
\normalsize

with $\mathcal{H}(\cdot)$ the entropy  is defined by : 

\begin{equation}\label{eq:MI2}
\mathcal{H}(P(y \mid \vx, \vomega))   = - \sum_{y}  P(y_i \mid \vx_i,\vomega)\log P(y_i \mid \vx_i,\vomega) 
\end{equation}

The good mutual information ratio highlights ABNN's superior ability to detect out-of-distribution samples compared to BNN. Moreover, Figure \ref{fig:diversity_tsne} illustrates that benefiting from multiple training instances, ABNN can effectively model multi-modes—a capability beyond the reach of classical BNNs. This is particularly valuable given the inherent multi-modal nature of the posterior.
\begin{table}[!htb]
    \renewcommand{\figurename}{Table}
    \centering
    \begin{tabular}{@{}lccc@{}}
        \toprule
        & ID & OOD & OOD/ID \\
        \midrule
        ABNN & 1.39e-4 &  5.22e-4 & \textbf{3.76} \\
        BNN  & 0.139 & 0.179 & 1.28 \\
        \bottomrule
    \end{tabular}
    \caption{\textbf{Evaluation of the average Mutual Information} on the test set and the OOD set of the different sample from an \method{} or a BNN. }\label{table:MutualINFO}
\end{table}


\section{Extra Experiments}
\label{sup:Experiments}
\subsection{ViT transfer learning}
\label{sup:ViT}
We also show that \method{} can be used in contexts of transfer learning. Table~\ref{table:transfer} presents the comparative performance of ViT B-16 pre-trained on ImageNet-21k and fine-tuned with and without a mono-modal \method{} on CIFAR-100 in 10,000 steps. Despite the mono-modality of \method{} in this experiment, we show that the corresponding ViT outperforms the classic fine-tuning in calibration and AUPR.

\begin{table}[!htb]
    \resizebox{\columnwidth}{!}
    {
    \centering
    \begin{tabular}{@{}lccccc@{}}
        \toprule
        & \textbf{Acc} & \textbf{ECE} & \textbf{AUPR} & \textbf{AUC} & \textbf{FPR95} \\
        \midrule
        Single model & \textbf{92.0} & 4.4 & 96.6 & \textbf{92.7} & \textbf{28.1} \\
        \method{} & 91.9 & \textbf{1.4} & \textbf{96.9} & 92.5 & 33.9 \\
        \bottomrule
    \end{tabular}
    }
    \caption{
     \textbf{Transfer learning of a ViT pre-trained on ImageNet-21k on CIFAR-100.} The first line is the classic pre-trained model fine-tuned with \method{}, and the second is fine-tuned with the classic layer normalization.
    }
    \label{table:transfer}
\end{table}

\subsection{Extra baselines}
\label{sup:baselines}

To benchmark our method against other posthoc techniques, we 
we implement a variant of Test-Time Augmentation~\cite{ashukha2019pitfalls,shanmugam2021better,son2023efficient}, incorporating random Gaussian noise with a specified standard deviation of $0.08$. The objective is to introduce diversity and ensemble different predictions by leveraging the added noise. Like \cite{lambert2018deep}, we experimented with adding noise to the latent space, representing the scenario where ABNN is not trained. We tested various levels of standard deviation, and the corresponding results are summarized in Table \ref{tab:noise}. Notably, the untrained ABNN does not perform effectively, underscoring the significance of a brief finetuning phase. Additionally, we train a VI-BNN, a non-posthoc technique, to understand the performance of a traditional BNN. It's noteworthy that VI-BNNs proved challenging to train and demonstrated subpar performance.

\begin{table*}[!htb]
    \centering
    \resizebox{0.6\textwidth}{!}
    {
    \begin{tabular}{@{}llccccc<{\kern-\tabcolsep}}
    \toprule
        & & \textbf{Acc} $\uparrow$ & \textbf{ECE} $\downarrow$ & \textbf{AUPR} $\uparrow$ & \textbf{AUC} $\uparrow$ & \textbf{FPR95} $\downarrow$  \\
        \midrule
        \multirow{8}{*}{\rotatebox[origin=c]{90}{\textbf{CIFAR-10}}} & Single model & \first \textbf{95.53} & \first \textbf{0.83} & \second 96.52 & \second 93.70 & \second 18.43\\
        & VI BNN & 75.66 & 5.40 & 80.60 & 69.53 & 66.62\\
        &  Test-Time Augmentation & 89.95 & 2.49 & 93.78 & 89.95 & 25.12 \\
         & Noise on latent space std=0.01 & \second 95.51 & \second 0.82 & 96.51 & 93.68 & 18.56 \\ 
        & Noise on latent space std=0.1 & 95.46 & 0.90 & 95.88 & 92.64 & 20.94 \\
        & Noise on latent space std=1 & 18.66 & 31.58 & 71.89 & 50.86 & 93.27 \\
        & \method{} & 95.43 & 0.85 & \first \textbf{97.03} & \first \textbf{94.73} & \first \textbf{15.11} \\
        
        \midrule
        
        \multirow{8}{*}{\rotatebox[origin=c]{90}{\textbf{CIFAR-100}}} &  Single model  & \first \textbf{79.05} & \first \textbf{5.34} & 88.72 & 79.96 & \second 52.04 \\
        & VI BNN & 41.17 & 8.97 & 78.44 & 61.82 & 86.15 \\
        & Test-Time Augmentation & 72.65 & 7.67 & 86.64 & 76.89 & 55.52 \\
        & Noise on latent space std=0.01 & \second 79.03 & \second 5.39 & \second 88.68 & \second 79.85 & 52.54 \\
        & Noise on latent space std=0.1 & 77.49 & 6.36 & 86.53 & 75.19 & 64.42 \\
        & Noise on latent space std=1 & 01.03 & 11.91 & 74.08 & 50.92 & 95.74 \\
        & \method{} & 78.94 & 5.47 & \first \textbf{89.36} & \first \textbf{81.04} & \first \textbf{50.12} \\

        \bottomrule
    \end{tabular}
    }
    \caption{
        \textbf{Performance comparison of different Post-hoc and BNN uncertainty quantification baselines on CIFAR-10/100 using ResNet-50.}
    } \label{tab:noise}
\end{table*}

\section{Training hyperparameters}
\label{sec:Implementation}

Table~\ref{table:Implementationdetails} provides a 
detailed overview of all the hyperparameters employed throughout our study. We use SGD in conjunction with a multistep learning-rate scheduler for image classification tasks, adjusting the rate by multiplying it by $\gamma$-lr at each milestone. It's important to note that, for stability reasons, BatchEnsemble based on ResNet-50 employed a lower learning rate of 0.08, deviating from the default 0.1. Our "Medium" data augmentation strategy encompasses a blend of Mixup~\citep{zhang2018mixup} and Cutmix~\citep{yun2019cutmix}, with a switch probability of 0.5. Additionally, 
\texttt{timm}'s augmentation classes~\citep{rw2019timm} were incorporated with coefficients of 0.5 and 0.2. RandAugment~\citep{cubuk2020randaugment} with parameters $m=9$, $n=2$, and $mstd=1$, along with label-smoothing~\citep{szegedy2016rethinking} of intensity 0.1, were also applied.

In the case of ImageNet, we follow the A3 procedure outlined in \cite{wightman2021resnet} for all models. It's worth mentioning that training according to the exact A3 procedure was not consistently feasible; please refer to the specific subsections for additional details.

We highlight that, to enhance training stability and fasten the training, we introduc a hyperparameter $\alpha$ in the BNL layer. This transforms the layer 
as follows:

\begin{align}\label{eq:BNL1_v2}
\BNL(\vh_j)  = \frac{\vh_j - \hat{\mu}_{j}}{\hat{\sigma}_{j} }\times\gamma_j (1+\vepsilon_{j}\alpha)+\beta_{j}.\\
\end{align}

The hyperparameter $\alpha$ is typically set to 0.01, except in the case of ViT, where specific considerations may apply.

\begin{table*}[t]

    \centering
    \resizebox{\textwidth}{!}
    {
    \begin{tabular}{@{}llcccccccc@{}}
        \toprule
        \textbf{Dataset} & \textbf{Networks} & Epochs & Batch size & start lr & Momentum & Weight decay & $\gamma$-lr & Milestones & Data augmentations \\
        \midrule
        C10 & R50 & 200 & 128 & 0.1 & 0.9 & 5$\cdot10^{-4}$ & 0.2 & 60, 120, 160 & HFlip\\
        C10 & WR28-10 & 200 & 128 & 0.1 & 0.9 & 5$\cdot10^{-4}$ & 0.2 & 60, 120, 160 & HFlip\\
        C100 & R50 & 200 & 128 & 0.1 & 0.9 & 5$\cdot10^{-4}$ & 0.2 & 60, 120, 160 & HFlip\\
        C100 & WR28-10 & 200 & 128 & 0.1 & 0.9 & 5$\cdot10^{-4}$ & 0.2 & 60, 120, 160 & Medium \\
        \bottomrule
    \end{tabular}
     }
    \caption{\textbf{Hyperparameters for image classification experiments.} HFlip denotes the classical horizontal flip.}
    \label{table:Implementationdetails}
\end{table*}

\begin{table*}[t]

    \centering
    \resizebox{\textwidth}{!}
    {
    \begin{tabular}{@{}llccccccccc@{}}
        \toprule
        \textbf{Dataset} & \textbf{Networks} & Epochs & Batch size & start lr & Alpha & Momentum & Weight decay & $\gamma$-lr & Milestones & Data augmentations \\
        \midrule
        C10 & R50 & 2 & 128 & 0,0057 & 0.01 & 0.9 & 5$\cdot10^{-4}$ & / & / & HFlip\\
        C10 & WR28-10 & 2 & 128 & 0,0091 & 0.01 & 0.9 & 5$\cdot10^{-4}$ & / & / & HFlip\\
        C100 & R50 & 10 & 128 & 0,00139 & 0.01 & 0.9 & 5$\cdot10^{-4}$ & 0.5 & 5 & HFlip\\
        C100 & WR28-10 & 10 & 128 & 0,034 & 0.01 & 0.9 & 5$\cdot10^{-4}$ & 0.5 & 5 & HFlip \\
        Imagenet & R50 & 1 & 128 & 0,00439 & 0,01 & 0.9 & 5$\cdot10^{-4}$ & / & / & HFlip \\
        Imagenet & ViT & 0.25 & 128 & 7.33$\cdot10^{-6}$ & 0.00035 & 0.9 & $7\cdot10^{-6}$ & $5\cdot10^{-4}$ & Constant & HFlip \\
        StreetHazards & DeepLabv3+ & 10 & 4 & 0.01 & 0.01 & 0.9 & 1$\cdot10^{-4}$ & 0.9 & Polynomial & HFlip, RandomCrop, ColorJitter, RandomScale \\
        BDD-Anomaly & DeepLabv3+ & 10 & 4 & 0.01 & 0.01 & 0.9 & 1$\cdot10^{-4}$ & 0.9 & / & HFlip, RandomCrop, ColorJitter, RandomScale \\
        MUAD & DeepLabv3+ & 6 & 4 & 0.044 & 0.01 & 0.9 & 1$\cdot10^{-4}$ & 0.9 & / & HFlip, RandomCrop, ColorJitter, RandomScale \\
        \bottomrule
    \end{tabular}
    }
    \caption{\textbf{Hyperparameters for image classification experiments with \method{}.} HFlip denotes the classical horizontal flip. Random prior has been used.}
    \label{table:ImplementationdetailsABNN}
\end{table*}

\input{sec/S1_notations}

%% file: sec/S1_notations.tex
\section{Notations}\label{sup:Notations}
We summarize the main notations used in the paper in Table \ref{table:notations}.
\begin{table*}[!htb]
\renewcommand{\figurename}{Table}
\caption{\textbf{Summary of the main notations of the paper.}}
\label{table:notations}
\centering
\resizebox{\textwidth}{!}
{
\begin{tabular}{@{}ll@{}}
\toprule
\textbf{Notations}                      & \textbf{Meaning}  \\ 
\midrule
$\mathcal{D}=\{ (\vx_i,\vy_i )\}_{i=1}^{N}$ & The set of $N$ data samples and the corresponding labels \\ 
$j$ & The index of the current layer\\
$\vomega$& The set of all the weights of the DNN  \\
$\vomega_{m} \sim P(\vomega|\mathcal{D})$& The $m$-th sample from the concatenation of weights of the posterior of the DNN.  \\
$M$ & The number of networks in an ensemble \\ 
$\vomega^{(t)}$& The concatenation of all the weights of the DNN after $t$ steps of optimization \\
$\vh_j$ & The pre-activation feature map and output of layer $(j-1)$ \& input of layer $j$  before  normalization\\
$\vu_j$ & The pre-activation feature map and output of layer $(j-1)$ \& input of layer $j$  before  normalization\\
$\gamma_j$ $\beta_j$  & The parameters of the batch, instance, or layer normalization of layer $j$ \\
$\mu_j$ $\sigma_j$  & The empirical mean and variance used by the batch, instance, or layer normalization of layer $j$ \\
$\va_j$ & The feature map and output of layer $j$, $\va_j = a (\vu_j)$ \\ 
$a(\cdot)$& The activation function  \\
\midrule
$W^{(j)}$& The weights of the $j$-th layer in a Multi-Layer Perceptron (MLP).\\
$W^{(j)}_{\mu}$& The mean weights of the $j$-th layer in a BNN MLP\\
$W^{(j)}_{\sigma}$& The standard variation weights of the $j$-th layer in a BNN MLP\\
$\epsilon^{(j)} \sim \mathcal{N}(\mathbf{0}, \mathds{1})$ & A vector sampled from a standard normal distribution %
at layer $j$\\
$\epsilon$ & The concatenation of all the $j-$ the $\epsilon^{(j)} $ of  each layer $j$\\
$\epsilon_l$ &  The $l$-th sample of $\epsilon$\\
$\mathcal{H}$ & The entropy function\\ 
\bottomrule
\end{tabular}
}

\end{table*}

%% file: main.bbl
\begin{thebibliography}{96}
\providecommand{\natexlab}[1]{#1}
\providecommand{\url}[1]{\texttt{#1}}
\expandafter\ifx\csname urlstyle\endcsname\relax
  \providecommand{\doi}[1]{doi: #1}\else
  \providecommand{\doi}{doi: \begingroup \urlstyle{rm}\Url}\fi

\bibitem[Arrieta et~al.(2020)Arrieta, D{\'\i}az-Rodr{\'\i}guez, Del~Ser, Bennetot, Tabik, Barbado, Garc{\'\i}a, Gil-L{\'o}pez, Molina, Benjamins, et~al.]{arrieta2020explainable}
Alejandro~Barredo Arrieta, Natalia D{\'\i}az-Rodr{\'\i}guez, Javier Del~Ser, Adrien Bennetot, Siham Tabik, Alberto Barbado, Salvador Garc{\'\i}a, Sergio Gil-L{\'o}pez, Daniel Molina, Richard Benjamins, et~al.
\newblock Explainable artificial intelligence (xai): Concepts, taxonomies, opportunities and challenges toward responsible ai.
\newblock \emph{Information fusion}, 2020.

\bibitem[Ashukha et~al.(2019)Ashukha, Lyzhov, Molchanov, and Vetrov]{ashukha2019pitfalls}
Arsenii Ashukha, Alexander Lyzhov, Dmitry Molchanov, and Dmitry Vetrov.
\newblock Pitfalls of in-domain uncertainty estimation and ensembling in deep learning.
\newblock In \emph{ICLR}, 2019.

\bibitem[Ba et~al.(2016)Ba, Kiros, and Hinton]{ba2016layer}
Jimmy~Lei Ba, Jamie~Ryan Kiros, and Geoffrey~E Hinton.
\newblock Layer normalization.
\newblock In \emph{NeurIPSW}, 2016.

\bibitem[Blei et~al.(2017)Blei, Kucukelbir, and McAuliffe]{blei2017variational}
David~M Blei, Alp Kucukelbir, and Jon~D McAuliffe.
\newblock Variational inference: A review for statisticians.
\newblock \emph{JASA}, 2017.

\bibitem[Blundell et~al.(2015)Blundell, Cornebise, Kavukcuoglu, and Wierstra]{blundell2015weight}
Charles Blundell, Julien Cornebise, Koray Kavukcuoglu, and Daan Wierstra.
\newblock Weight uncertainty in neural network.
\newblock In \emph{ICML}, 2015.

\bibitem[Brosse et~al.(2020)Brosse, Riquelme, Martin, Gelly, and Moulines]{brosse2020last}
Nicolas Brosse, Carlos Riquelme, Alice Martin, Sylvain Gelly, and {\'E}ric Moulines.
\newblock On last-layer algorithms for classification: Decoupling representation from uncertainty estimation.
\newblock \emph{arXiv preprint arXiv:2001.08049}, 2020.

\bibitem[Chan et~al.(2021)Chan, Lis, Uhlemeyer, Blum, Honari, Siegwart, Fua, Salzmann, and Rottmann]{chan2021segmentmeifyoucan}
Robin Chan, Krzysztof Lis, Svenja Uhlemeyer, Hermann Blum, Sina Honari, Roland Siegwart, Pascal Fua, Mathieu Salzmann, and Matthias Rottmann.
\newblock Segmentmeifyoucan: A benchmark for anomaly segmentation.
\newblock In \emph{NeurIPS}, 2021.

\bibitem[Chen et~al.(2018)Chen, Zhu, Papandreou, Schroff, and Adam]{chen2018encoder}
Liang-Chieh Chen, Yukun Zhu, George Papandreou, Florian Schroff, and Hartwig Adam.
\newblock Encoder-decoder with atrous separable convolution for semantic image segmentation.
\newblock In \emph{ECCV}, 2018.

\bibitem[Corbi{\`e}re et~al.(2021)Corbi{\`e}re, Lafon, Thome, Cord, and P{\'e}rez]{corbiere2021beyond}
Charles Corbi{\`e}re, Marc Lafon, Nicolas Thome, Matthieu Cord, and Patrick P{\'e}rez.
\newblock Beyond first-order uncertainty estimation with evidential models for open-world recognition.
\newblock In \emph{ICMLW}, 2021.

\bibitem[Cordts et~al.(2016)Cordts, Omran, Ramos, Rehfeld, Enzweiler, Benenson, Franke, Roth, and Schiele]{cordts2016cityscapes}
Marius Cordts, Mohamed Omran, Sebastian Ramos, Timo Rehfeld, Markus Enzweiler, Rodrigo Benenson, Uwe Franke, Stefan Roth, and Bernt Schiele.
\newblock The cityscapes dataset for semantic urban scene understanding.
\newblock In \emph{CVPR}, 2016.

\bibitem[Cubuk et~al.(2020)Cubuk, Zoph, Shlens, and Le]{cubuk2020randaugment}
Ekin~D Cubuk, Barret Zoph, Jonathon Shlens, and Quoc~V Le.
\newblock Randaugment: Practical automated data augmentation with a reduced search space.
\newblock In \emph{CVPR}, 2020.

\bibitem[Daxberger et~al.(2021{\natexlab{a}})Daxberger, Kristiadi, Immer, Eschenhagen, Bauer, and Hennig]{laplace2021}
Erik Daxberger, Agustinus Kristiadi, Alexander Immer, Runa Eschenhagen, Matthias Bauer, and Philipp Hennig.
\newblock Laplace redux--effortless {B}ayesian deep learning.
\newblock In \emph{{N}eur{IPS}}, 2021{\natexlab{a}}.

\bibitem[Daxberger et~al.(2021{\natexlab{b}})Daxberger, Nalisnick, Allingham, Antor{\'a}n, and Hern{\'a}ndez-Lobato]{daxberger2021bayesian}
Erik Daxberger, Eric Nalisnick, James~U Allingham, Javier Antor{\'a}n, and Jos{\'e}~Miguel Hern{\'a}ndez-Lobato.
\newblock Bayesian deep learning via subnetwork inference.
\newblock In \emph{ICML}, 2021{\natexlab{b}}.

\bibitem[Dehghani et~al.(2023)Dehghani, Djolonga, Mustafa, Padlewski, Heek, Gilmer, Steiner, Caron, Geirhos, Alabdulmohsin, et~al.]{dehghani2023scaling}
Mostafa Dehghani, Josip Djolonga, Basil Mustafa, Piotr Padlewski, Jonathan Heek, Justin Gilmer, Andreas~Peter Steiner, Mathilde Caron, Robert Geirhos, Ibrahim Alabdulmohsin, et~al.
\newblock Scaling vision transformers to 22 billion parameters.
\newblock In \emph{ICML}, 2023.

\bibitem[Deng et~al.(2009)Deng, Dong, Socher, Li, Li, and Fei-Fei]{deng2009imagenet}
Jia Deng, Wei Dong, Richard Socher, Li-Jia Li, Kai Li, and Li Fei-Fei.
\newblock Imagenet: A large-scale hierarchical image database.
\newblock In \emph{CVPR}, 2009.

\bibitem[Depeweg et~al.(2018)Depeweg, Hernandez-Lobato, Doshi-Velez, and Udluft]{depeweg2018decomposition}
Stefan Depeweg, Jose-Miguel Hernandez-Lobato, Finale Doshi-Velez, and Steffen Udluft.
\newblock Decomposition of uncertainty in bayesian deep learning for efficient and risk-sensitive learning.
\newblock In \emph{ICML}, 2018.

\bibitem[Devlin et~al.(2018)Devlin, Chang, Lee, and Toutanova]{devlin2018bert}
Jacob Devlin, Ming-Wei Chang, Kenton Lee, and Kristina Toutanova.
\newblock Bert: Pre-training of deep bidirectional transformers for language understanding.
\newblock \emph{arXiv preprint arXiv:1810.04805}, 2018.

\bibitem[Dosovitskiy et~al.(2021)Dosovitskiy, Beyer, Kolesnikov, Weissenborn, Zhai, Unterthiner, Dehghani, Minderer, Heigold, Gelly, et~al.]{dosovitskiy2020image}
Alexey Dosovitskiy, Lucas Beyer, Alexander Kolesnikov, Dirk Weissenborn, Xiaohua Zhai, Thomas Unterthiner, Mostafa Dehghani, Matthias Minderer, Georg Heigold, Sylvain Gelly, et~al.
\newblock An image is worth 16x16 words: Transformers for image recognition at scale.
\newblock In \emph{ICLR}, 2021.

\bibitem[Durasov et~al.(2021)Durasov, Bagautdinov, Baque, and Fua]{durasov2021masksembles}
Nikita Durasov, Timur Bagautdinov, Pierre Baque, and Pascal Fua.
\newblock Masksembles for uncertainty estimation.
\newblock In \emph{CVPR}, 2021.

\bibitem[Dusenberry et~al.(2020)Dusenberry, Jerfel, Wen, Ma, Snoek, Heller, Lakshminarayanan, and Tran]{dusenberry2020efficient}
Michael Dusenberry, Ghassen Jerfel, Yeming Wen, Yian Ma, Jasper Snoek, Katherine Heller, Balaji Lakshminarayanan, and Dustin Tran.
\newblock Efficient and scalable bayesian neural nets with rank-1 factors.
\newblock In \emph{ICML}, 2020.

\bibitem[Feller(1991)]{feller1991introduction}
William Feller.
\newblock \emph{An introduction to probability theory and its applications, Volume 2}.
\newblock John Wiley \& Sons, 1991.

\bibitem[Fort et~al.(2019)Fort, Hu, and Lakshminarayanan]{fort2019deep}
Stanislav Fort, Huiyi Hu, and Balaji Lakshminarayanan.
\newblock Deep ensembles: A loss landscape perspective.
\newblock \emph{arXiv preprint arXiv:1912.02757}, 2019.

\bibitem[Franchi et~al.(2020)Franchi, Bursuc, Aldea, Dubuisson, and Bloch]{franchi2020tradi}
Gianni Franchi, Andrei Bursuc, Emanuel Aldea, S{\'e}verine Dubuisson, and Isabelle Bloch.
\newblock Tradi: Tracking deep neural network weight distributions.
\newblock In \emph{ECCV}, 2020.

\bibitem[Franchi et~al.(2022)Franchi, Yu, Bursuc, Tena, Kazmierczak, Dubuisson, Aldea, and Filliat]{franchi2022muad}
Gianni Franchi, Xuanlong Yu, Andrei Bursuc, Angel Tena, R{\'e}mi Kazmierczak, S{\'e}verine Dubuisson, Emanuel Aldea, and David Filliat.
\newblock Muad: Multiple uncertainties for autonomous driving, a benchmark for multiple uncertainty types and tasks.
\newblock In \emph{BMVC}, 2022.

\bibitem[Franchi et~al.(2023)Franchi, Bursuc, Aldea, Dubuisson, and Bloch]{franchi2020encoding}
Gianni Franchi, Andrei Bursuc, Emanuel Aldea, S{\'e}verine Dubuisson, and Isabelle Bloch.
\newblock Encoding the latent posterior of bayesian neural networks for uncertainty quantification.
\newblock \emph{T-PAMI}, 2023.

\bibitem[Gal and Ghahramani(2016)]{gal2016dropout}
Yarin Gal and Zoubin Ghahramani.
\newblock Dropout as a bayesian approximation: Representing model uncertainty in deep learning.
\newblock In \emph{ICML}, 2016.

\bibitem[Gawlikowski et~al.(2023)Gawlikowski, Tassi, Ali, Lee, Humt, Feng, Kruspe, Triebel, Jung, Roscher, et~al.]{gawlikowski2023survey}
Jakob Gawlikowski, Cedrique Rovile~Njieutcheu Tassi, Mohsin Ali, Jongseok Lee, Matthias Humt, Jianxiang Feng, Anna Kruspe, Rudolph Triebel, Peter Jung, Ribana Roscher, et~al.
\newblock A survey of uncertainty in deep neural networks.
\newblock \emph{Artificial Intelligence Review}, 2023.

\bibitem[Goan and Fookes(2020)]{goan2020bayesian}
Ethan Goan and Clinton Fookes.
\newblock Bayesian neural networks: An introduction and survey.
\newblock \emph{Case Studies in Applied Bayesian Data Science}, 2020.

\bibitem[Hansen and Salamon(1990)]{hansen1990neural}
Lars~Kai Hansen and Peter Salamon.
\newblock Neural network ensembles.
\newblock \emph{IEEE transactions on pattern analysis and machine intelligence}, 1990.

\bibitem[Havasi et~al.(2021)Havasi, Jenatton, Fort, Liu, Snoek, Lakshminarayanan, Dai, and Tran]{havasi2021training}
Marton Havasi, Rodolphe Jenatton, Stanislav Fort, Jeremiah~Zhe Liu, Jasper Snoek, Balaji Lakshminarayanan, Andrew~Mingbo Dai, and Dustin Tran.
\newblock Training independent subnetworks for robust prediction.
\newblock In \emph{ICLR}, 2021.

\bibitem[He et~al.(2016)He, Zhang, Ren, and Sun]{he2016deep}
Kaiming He, Xiangyu Zhang, Shaoqing Ren, and Jian Sun.
\newblock Deep residual learning for image recognition.
\newblock In \emph{CVPR}, 2016.

\bibitem[He et~al.(2017)He, Gkioxari, Doll{\'a}r, and Girshick]{he2017mask}
Kaiming He, Georgia Gkioxari, Piotr Doll{\'a}r, and Ross Girshick.
\newblock Mask r-cnn.
\newblock In \emph{ICCV}, 2017.

\bibitem[Hein et~al.(2019)Hein, Andriushchenko, and Bitterwolf]{hein2019relu}
Matthias Hein, Maksym Andriushchenko, and Julian Bitterwolf.
\newblock Why relu networks yield high-confidence predictions far away from the training data and how to mitigate the problem.
\newblock In \emph{CVPR}, 2019.

\bibitem[Hendrycks and Gimpel(2017)]{hendrycks2016baseline}
Dan Hendrycks and Kevin Gimpel.
\newblock A baseline for detecting misclassified and out-of-distribution examples in neural networks.
\newblock In \emph{ICLR}, 2017.

\bibitem[Hendrycks et~al.(2019)Hendrycks, Basart, Mazeika, Mostajabi, Steinhardt, and Song]{hendrycks2019anomalyseg}
Dan Hendrycks, Steven Basart, Mantas Mazeika, Mohammadreza Mostajabi, Jacob Steinhardt, and Dawn Song.
\newblock A benchmark for anomaly segmentation.
\newblock \emph{arXiv preprint arXiv:1911.11132}, 2019.

\bibitem[Hendrycks et~al.(2021{\natexlab{a}})Hendrycks, Basart, Mu, Kadavath, Wang, Dorundo, Desai, Zhu, Parajuli, Guo, et~al.]{hendrycks2021jacob}
Dan Hendrycks, Steven Basart, Norman Mu, Saurav Kadavath, Frank Wang, Evan Dorundo, Rahul Desai, Tyler Zhu, Samyak Parajuli, Mike Guo, et~al.
\newblock Jacob steinhardt et justin gilmer. the many faces of robustness: A critical analysis of out-of-distribution generalization.
\newblock In \emph{ICCV}, 2021{\natexlab{a}}.

\bibitem[Hendrycks et~al.(2021{\natexlab{b}})Hendrycks, Carlini, Schulman, and Steinhardt]{hendrycks2021unsolved}
Dan Hendrycks, Nicholas Carlini, John Schulman, and Jacob Steinhardt.
\newblock Unsolved problems in ml safety.
\newblock \emph{arXiv preprint arXiv:2109.13916}, 2021{\natexlab{b}}.

\bibitem[Hern{\'a}ndez-Lobato and Adams(2015)]{hernandez2015probabilistic}
Jos{\'e}~Miguel Hern{\'a}ndez-Lobato and Ryan Adams.
\newblock Probabilistic backpropagation for scalable learning of bayesian neural networks.
\newblock In \emph{ICML}, 2015.

\bibitem[Hora(1996)]{hora1996aleatory}
Stephen~C Hora.
\newblock Aleatory and epistemic uncertainty in probability elicitation with an example from hazardous waste management.
\newblock \emph{Reliability Engineering \& System Safety}, 1996.

\bibitem[Hron et~al.(2022)Hron, Novak, Pennington, and Sohl-Dickstein]{hron2022wide}
Jiri Hron, Roman Novak, Jeffrey Pennington, and Jascha Sohl-Dickstein.
\newblock Wide bayesian neural networks have a simple weight posterior: theory and accelerated sampling.
\newblock In \emph{ICML}, pages 8926--8945. PMLR, 2022.

\bibitem[H{\"u}llermeier and Waegeman(2021)]{hullermeier2021aleatoric}
Eyke H{\"u}llermeier and Willem Waegeman.
\newblock Aleatoric and epistemic uncertainty in machine learning: An introduction to concepts and methods.
\newblock \emph{Machine Learning}, 2021.

\bibitem[Ilharco et~al.(2021)Ilharco, Wortsman, Wightman, Gordon, Carlini, Taori, Dave, Shankar, Namkoong, Miller, Hajishirzi, Farhadi, and Schmidt]{ilharco2021openclip}
Gabriel Ilharco, Mitchell Wortsman, Ross Wightman, Cade Gordon, Nicholas Carlini, Rohan Taori, Achal Dave, Vaishaal Shankar, Hongseok Namkoong, John Miller, Hannaneh Hajishirzi, Ali Farhadi, and Ludwig Schmidt.
\newblock Openclip, 2021.

\bibitem[Ioffe and Szegedy(2015)]{ioffe2015batch}
Sergey Ioffe and Christian Szegedy.
\newblock Batch normalization: Accelerating deep network training by reducing internal covariate shift.
\newblock In \emph{ICML}, 2015.

\bibitem[Izmailov et~al.(2021)Izmailov, Vikram, Hoffman, and Wilson]{izmailov2021bayesian}
Pavel Izmailov, Sharad Vikram, Matthew~D Hoffman, and Andrew Gordon~Gordon Wilson.
\newblock What are bayesian neural network posteriors really like?
\newblock In \emph{ICML}, 2021.

\bibitem[Jordan et~al.(1999)Jordan, Ghahramani, Jaakkola, and Saul]{jordan1999introduction}
Michael~I Jordan, Zoubin Ghahramani, Tommi~S Jaakkola, and Lawrence~K Saul.
\newblock An introduction to variational methods for graphical models.
\newblock \emph{ML}, 1999.

\bibitem[Kendall and Gal(2017)]{kendall2017uncertainties}
Alex Kendall and Yarin Gal.
\newblock What uncertainties do we need in bayesian deep learning for computer vision?
\newblock \emph{NeurIPS}, 2017.

\bibitem[Kirillov et~al.(2023)Kirillov, Mintun, Ravi, Mao, Rolland, Gustafson, Xiao, Whitehead, Berg, Lo, et~al.]{kirillov2023segment}
Alexander Kirillov, Eric Mintun, Nikhila Ravi, Hanzi Mao, Chloe Rolland, Laura Gustafson, Tete Xiao, Spencer Whitehead, Alexander~C Berg, Wan-Yen Lo, et~al.
\newblock Segment anything.
\newblock \emph{arXiv preprint arXiv:2304.02643}, 2023.

\bibitem[Kristiadi et~al.(2020)Kristiadi, Hein, and Hennig]{kristiadi2020being}
Agustinus Kristiadi, Matthias Hein, and Philipp Hennig.
\newblock Being bayesian, even just a bit, fixes overconfidence in relu networks.
\newblock In \emph{ICML}, 2020.

\bibitem[Krizhevsky(2009)]{krizhevsky2009learning}
Alex Krizhevsky.
\newblock Learning multiple layers of features from tiny images.
\newblock Technical report, MIT, 2009.

\bibitem[Krizhevsky et~al.(2012)Krizhevsky, Sutskever, and Hinton]{alexnet2012neurips}
A Krizhevsky, I Sutskever, and G Hinton.
\newblock Imagenet classification with deep convolutional networks.
\newblock In \emph{NeurIPS}, 2012.

\bibitem[Lakshminarayanan et~al.(2017)Lakshminarayanan, Pritzel, and Blundell]{lakshminarayanan2017simple}
Balaji Lakshminarayanan, Alexander Pritzel, and Charles Blundell.
\newblock Simple and scalable predictive uncertainty estimation using deep ensembles.
\newblock In \emph{NeurIPS}, 2017.

\bibitem[Lambert et~al.(2018)Lambert, Sener, and Savarese]{lambert2018deep}
John Lambert, Ozan Sener, and Silvio Savarese.
\newblock Deep learning under privileged information using heteroscedastic dropout.
\newblock In \emph{CVPR}, pages 8886--8895, 2018.

\bibitem[Laurent et~al.(2023{\natexlab{a}})Laurent, Aldea, and Franchi]{laurent2023symmetry}
Olivier Laurent, Emanuel Aldea, and Gianni Franchi.
\newblock A symmetry-aware exploration of bayesian neural network posteriors.
\newblock \emph{arXiv preprint arXiv:2310.08287}, 2023{\natexlab{a}}.

\bibitem[Laurent et~al.(2023{\natexlab{b}})Laurent, Lafage, Tartaglione, Daniel, Martinez, Bursuc, and Franchi]{laurent2023packed}
Olivier Laurent, Adrien Lafage, Enzo Tartaglione, Geoffrey Daniel, Jean-Marc Martinez, Andrei Bursuc, and Gianni Franchi.
\newblock Packed-ensembles for efficient uncertainty estimation.
\newblock In \emph{ICLR}, 2023{\natexlab{b}}.

\bibitem[Lee et~al.(2018)Lee, Lee, Lee, and Shin]{lee2018simple}
Kimin Lee, Kibok Lee, Honglak Lee, and Jinwoo Shin.
\newblock A simple unified framework for detecting out-of-distribution samples and adversarial attacks.
\newblock In \emph{NeurIPS}, 2018.

\bibitem[Li et~al.(2022)Li, Chen, Wang, Hong, Ye, Han, Chen, Zhang, Xu, Yeung, et~al.]{li2022coda}
Kaican Li, Kai Chen, Haoyu Wang, Lanqing Hong, Chaoqiang Ye, Jianhua Han, Yukuai Chen, Wei Zhang, Chunjing Xu, Dit-Yan Yeung, et~al.
\newblock Coda: A real-world road corner case dataset for object detection in autonomous driving.
\newblock In \emph{ECCV}, 2022.

\bibitem[Liu et~al.(2022)Liu, Mao, Wu, Feichtenhofer, Darrell, and Xie]{liu2022convnet}
Zhuang Liu, Hanzi Mao, Chao-Yuan Wu, Christoph Feichtenhofer, Trevor Darrell, and Saining Xie.
\newblock A convnet for the 2020s.
\newblock In \emph{CVPR}, 2022.

\bibitem[MacKay(1992)]{mackay1992practical}
David~JC MacKay.
\newblock A practical bayesian framework for backpropagation networks.
\newblock \emph{Neural computation}, 1992.

\bibitem[Maddox et~al.(2019)Maddox, Izmailov, Garipov, Vetrov, and Wilson]{maddox2019simple}
Wesley~J Maddox, Pavel Izmailov, Timur Garipov, Dmitry~P Vetrov, and Andrew~Gordon Wilson.
\newblock A simple baseline for bayesian uncertainty in deep learning.
\newblock In \emph{NeurIPS}, 2019.

\bibitem[Malinin and Gales(2018)]{malinin2018predictive}
Andrey Malinin and Mark Gales.
\newblock Predictive uncertainty estimation via prior networks.
\newblock In \emph{NeurIPS}, 2018.

\bibitem[Naeini et~al.(2015)Naeini, Cooper, and Hauskrecht]{naeini2015obtaining}
Mahdi~Pakdaman Naeini, Gregory~F. Cooper, and Milos Hauskrecht.
\newblock Obtaining well calibrated probabilities using bayesian binning.
\newblock In \emph{AAAI}, 2015.

\bibitem[Nalisnick(2018)]{nalisnick2018priors}
Eric~Thomas Nalisnick.
\newblock \emph{On priors for Bayesian neural networks}.
\newblock University of California, Irvine, 2018.

\bibitem[Nayman et~al.(2022)Nayman, Golbert, Noy, Ping, and Zelnik-Manor]{nayman2022diverse}
Niv Nayman, Avram Golbert, Asaf Noy, Tan Ping, and Lihi Zelnik-Manor.
\newblock Diverse imagenet models transfer better.
\newblock \emph{arXiv preprint arXiv:2204.09134}, 2022.

\bibitem[Neal(2012)]{neal2012bayesian}
Radford~M Neal.
\newblock \emph{Bayesian learning for neural networks}.
\newblock 2012.

\bibitem[Netzer et~al.(2011)Netzer, Wang, Coates, Bissacco, Wu, and Ng]{netzer2011reading}
Yuval Netzer, Tao Wang, Adam Coates, Alessandro Bissacco, Bo Wu, and Andrew~Y. Ng.
\newblock Reading digits in natural images with unsupervised feature learning.
\newblock In \emph{NeurIPSW}, 2011.

\bibitem[Neuhaus et~al.(2023)Neuhaus, Augustin, Boreiko, and Hein]{neuhaus2023spurious}
Yannic Neuhaus, Maximilian Augustin, Valentyn Boreiko, and Matthias Hein.
\newblock Spurious features everywhere-large-scale detection of harmful spurious features in imagenet.
\newblock In \emph{ICCV}, 2023.

\bibitem[Ovadia et~al.(2019)Ovadia, Fertig, Ren, Nado, Sculley, Nowozin, Dillon, Lakshminarayanan, and Snoek]{ovadia2019can}
Yaniv Ovadia, Emily Fertig, Jie Ren, Zachary Nado, David Sculley, Sebastian Nowozin, Joshua Dillon, Balaji Lakshminarayanan, and Jasper Snoek.
\newblock Can you trust your model's uncertainty? evaluating predictive uncertainty under dataset shift.
\newblock In \emph{NeurIPS}, 2019.

\bibitem[Paszke et~al.(2019)Paszke, Gross, Massa, Lerer, Bradbury, Chanan, Killeen, Lin, Gimelshein, Antiga, et~al.]{paszke2019pytorch}
Adam Paszke, Sam Gross, Francisco Massa, Adam Lerer, James Bradbury, Gregory Chanan, Trevor Killeen, Zeming Lin, Natalia Gimelshein, Luca Antiga, et~al.
\newblock Pytorch: An imperative style, high-performance deep learning library.
\newblock In \emph{NeurIPS}, 2019.

\bibitem[Radford et~al.(2018)Radford, Narasimhan, Salimans, Sutskever, et~al.]{radford2018improving}
Alec Radford, Karthik Narasimhan, Tim Salimans, Ilya Sutskever, et~al.
\newblock Improving language understanding by generative pre-training.
\newblock Technical report, OpenAI, 2018.

\bibitem[Redmon et~al.(2016)Redmon, Divvala, Girshick, and Farhadi]{redmon2016you}
Joseph Redmon, Santosh Divvala, Ross Girshick, and Ali Farhadi.
\newblock You only look once: Unified, real-time object detection.
\newblock In \emph{CVPR}, 2016.

\bibitem[Ridnik et~al.(2021)Ridnik, Ben-Baruch, Noy, and Zelnik-Manor]{ridnik2021imagenet21k}
Tal Ridnik, Emanuel Ben-Baruch, Asaf Noy, and Lihi Zelnik-Manor.
\newblock Imagenet-21k pretraining for the masses, 2021.

\bibitem[Ritter et~al.(2018)Ritter, Botev, and Barber]{ritter2018scalable}
Hippolyt Ritter, Aleksandar Botev, and David Barber.
\newblock A scalable laplace approximation for neural networks.
\newblock In \emph{ICLR}, 2018.

\bibitem[Rombach et~al.(2022)Rombach, Blattmann, Lorenz, Esser, and Ommer]{rombach2022high}
Robin Rombach, Andreas Blattmann, Dominik Lorenz, Patrick Esser, and Bj{\"o}rn Ommer.
\newblock High-resolution image synthesis with latent diffusion models.
\newblock In \emph{CVPR}, 2022.

\bibitem[Roy et~al.(2022)Roy, Trapp, Pilzer, Kannala, Sebe, Ricci, and Solin]{subhankar2022uncertainty}
Subhankar Roy, Martin Trapp, Andrea Pilzer, Juho Kannala, Nicu Sebe, Elisa Ricci, and Arno Solin.
\newblock Uncertainty-guided source-free domain adaptation.
\newblock In \emph{ECCV}, 2022.

\bibitem[Schuhmann et~al.(2022)Schuhmann, Beaumont, Vencu, Gordon, Wightman, Cherti, Coombes, Katta, Mullis, Wortsman, et~al.]{schuhmann2022laion}
Christoph Schuhmann, Romain Beaumont, Richard Vencu, Cade Gordon, Ross Wightman, Mehdi Cherti, Theo Coombes, Aarush Katta, Clayton Mullis, Mitchell Wortsman, et~al.
\newblock Laion-5b: An open large-scale dataset for training next generation image-text models.
\newblock \emph{NeurIPS}, 2022.

\bibitem[Severyn and Moschitti(2015)]{severyn2015}
Aliaksei Severyn and Alessandro Moschitti.
\newblock Learning to rank short text pairs with convolutional deep neural networks.
\newblock In \emph{SIGIR}, 2015.

\bibitem[Shanmugam et~al.(2021)Shanmugam, Blalock, Balakrishnan, and Guttag]{shanmugam2021better}
Divya Shanmugam, Davis Blalock, Guha Balakrishnan, and John Guttag.
\newblock Better aggregation in test-time augmentation.
\newblock In \emph{ICCV}, pages 1214--1223, 2021.

\bibitem[Son and Kang(2023)]{son2023efficient}
Jongwook Son and Seokho Kang.
\newblock Efficient improvement of classification accuracy via selective test-time augmentation.
\newblock \emph{Information Sciences}, 642:\penalty0 119148, 2023.

\bibitem[Szegedy et~al.(2016)Szegedy, Vanhoucke, Ioffe, Shlens, and Wojna]{szegedy2016rethinking}
Christian Szegedy, Vincent Vanhoucke, Sergey Ioffe, Jon Shlens, and Zbigniew Wojna.
\newblock Rethinking the inception architecture for computer vision.
\newblock In \emph{CVPR}, 2016.

\bibitem[Tierney and Kadane(1986)]{tierney1986accurate}
Luke Tierney and Joseph~B Kadane.
\newblock Accurate approximations for posterior moments and marginal densities.
\newblock \emph{Journal of the American Statistical Association}, 1986.

\bibitem[Tishby et~al.(1989)Tishby, Levin, and Solla]{tishby1989consistent}
Tishby, Levin, and Solla.
\newblock Consistent inference of probabilities in layered networks: predictions and generalizations.
\newblock In \emph{IJCNN}, 1989.

\bibitem[Tran et~al.(2022)Tran, Liu, Dusenberry, Phan, Collier, Ren, Han, Wang, Mariet, Hu, et~al.]{tran2022plex}
Dustin Tran, Jeremiah Liu, Michael~W Dusenberry, Du Phan, Mark Collier, Jie Ren, Kehang Han, Zi Wang, Zelda Mariet, Huiyi Hu, et~al.
\newblock Plex: Towards reliability using pretrained large model extensions.
\newblock \emph{arXiv preprint arXiv:2207.07411}, 2022.

\bibitem[Ulyanov et~al.(2016)Ulyanov, Vedaldi, and Lempitsky]{ulyanov2016instance}
Dmitry Ulyanov, Andrea Vedaldi, and Victor Lempitsky.
\newblock Instance normalization: The missing ingredient for fast stylization.
\newblock \emph{arXiv preprint arXiv:1607.08022}, 2016.

\bibitem[Wang et~al.(2022)Wang, Li, Feng, and Zhang]{wang2022vim}
Haoqi Wang, Zhizhong Li, Litong Feng, and Wayne Zhang.
\newblock {ViM}: Out-of-distribution with virtual-logit matching.
\newblock In \emph{CVPR}, 2022.

\bibitem[Welling and Teh(2011)]{welling2011bayesian}
Max Welling and Yee~W Teh.
\newblock Bayesian learning via stochastic gradient langevin dynamics.
\newblock In \emph{ICML}, 2011.

\bibitem[Wen et~al.(2019)Wen, Tran, and Ba]{wen2019batchensemble}
Yeming Wen, Dustin Tran, and Jimmy Ba.
\newblock {BatchEnsemble}: an alternative approach to efficient ensemble and lifelong learning.
\newblock In \emph{ICLR}, 2019.

\bibitem[Wightman(2019)]{rw2019timm}
Ross Wightman.
\newblock Pytorch image models.
\newblock \url{https://github.com/rwightman/pytorch-image-models}, 2019.

\bibitem[Wightman et~al.(2021)Wightman, Touvron, and Jegou]{wightman2021resnet}
Ross Wightman, Hugo Touvron, and Herve Jegou.
\newblock Resnet strikes back: An improved training procedure in timm.
\newblock In \emph{NeurIPSW}, 2021.

\bibitem[Wilson and Izmailov(2020)]{wilson2020bayesian}
Andrew~G Wilson and Pavel Izmailov.
\newblock Bayesian deep learning and a probabilistic perspective of generalization.
\newblock \emph{NeurIPS}, 2020.

\bibitem[Xia and Bouganis(2023)]{xia2023window}
Guoxuan Xia and Christos-Savvas Bouganis.
\newblock Window-based early-exit cascades for uncertainty estimation: When deep ensembles are more efficient than single models.
\newblock In \emph{ICCV}, 2023.

\bibitem[Yu et~al.(2020)Yu, Chen, Wang, Xian, Chen, Liu, Madhavan, and Darrell]{yu2020bdd100k}
Fisher Yu, Haofeng Chen, Xin Wang, Wenqi Xian, Yingying Chen, Fangchen Liu, Vashisht Madhavan, and Trevor Darrell.
\newblock Bdd100k: A diverse driving dataset for heterogeneous multitask learning.
\newblock In \emph{CVPR}, 2020.

\bibitem[Yun et~al.(2019)Yun, Han, Oh, Chun, Choe, and Yoo]{yun2019cutmix}
Sangdoo Yun, Dongyoon Han, Seong~Joon Oh, Sanghyuk Chun, Junsuk Choe, and Youngjoon Yoo.
\newblock Cutmix: Regularization strategy to train strong classifiers with localizable features.
\newblock In \emph{CVPR}, 2019.

\bibitem[Zablocki et~al.(2022)Zablocki, Ben-Younes, P{\'e}rez, and Cord]{zablocki2022explainability}
{\'E}loi Zablocki, H{\'e}di Ben-Younes, Patrick P{\'e}rez, and Matthieu Cord.
\newblock Explainability of deep vision-based autonomous driving systems: Review and challenges.
\newblock \emph{IJCV}, 2022.

\bibitem[Zagoruyko and Komodakis(2016)]{zagoruyko2016wide}
Sergey Zagoruyko and Nikos Komodakis.
\newblock Wide residual networks.
\newblock In \emph{BMVC}, 2016.

\bibitem[Zhai et~al.(2023)Zhai, Mustafa, Kolesnikov, and Beyer]{zhai2023sigmoid}
Xiaohua Zhai, Basil Mustafa, Alexander Kolesnikov, and Lucas Beyer.
\newblock Sigmoid loss for language image pre-training.
\newblock In \emph{ICCV}, 2023.

\bibitem[Zhang et~al.(2018)Zhang, Cisse, Dauphin, and Lopez-Paz]{zhang2018mixup}
Hongyi Zhang, Moustapha Cisse, Yann~N Dauphin, and David Lopez-Paz.
\newblock mixup: Beyond empirical risk minimization.
\newblock In \emph{ICLR}, 2018.

\end{thebibliography}
